\documentclass{article}

\usepackage{microtype}
\usepackage{graphicx}
\usepackage{subfigure}
\usepackage{booktabs} % for professional tables

\usepackage{hyperref}

\usepackage[accepted]{icml2020}
\usepackage{xspace}

\usepackage{overpic}
\usepackage{caption}        % captionof
\usepackage{amsthm}
\usepackage{bbm}
\usepackage{amsmath}
\usepackage{mathtools}
\usepackage{amssymb}
\usepackage{float}

\usepackage{pgfplots, pgfplotstable}
\usepackage{algorithm}

\usepackage{enumitem}
\usepackage{multirow}
\usepackage[title]{appendix}
\usepackage{listings}
\usepackage{xcolor}
\usepackage{wrapfig}

\definecolor{dkgreen}{rgb}{0,0.6,0}
\definecolor{gray}{rgb}{0.5,0.5,0.5}
\definecolor{mauve}{rgb}{0.58,0,0.82}

\newcommand{\AT}[1]{\textcolor{purple}{[AT: {#1}]}} % Andrea
\newcommand{\david}[1]{\textcolor{red}{[DF: {#1}]}} % David
\definecolor{Fcolor}{HTML}{af2418}

\definecolor{Ccolor}{HTML}{ffd359}

\def\Real{{\mathbb R}}

\def\int{\mathrm{int}}

\newcommand{\todo}[1]{\textcolor{red}{#1}}  % To Do
\newcommand{\Figure}[1]{Figure~\ref{fig:#1}}

\newcommand{\Table}[1]{Table~\ref{tab:#1}}
\newcommand{\eq}[1]{(\ref{eq:#1})}

\renewcommand{\paragraph}[1]{\vspace{.1em}\noindent\textbf{#1.~}}

\usepackage{lipsum}

\def\image{\mathbf{I}}
\renewcommand{\next}[1]{{#1}'}
\def\params{\omega}
\def\encoder{\mathcal{E}_\params}

\def\masknet{\mathcal{D}_\params}

\def\capsule{\mathbf{c}}
\def\pose{\boldsymbol{\theta}}
\def\flow{\Phi}
\def\flowgt{\Phi_\text{gt}}

\newcommand{\depth}{d}

\newcommand{\shape}{{\bf s}}

\newcommand{\pixel}{{\bf u}} % pixel coordinates

\newcommand{\expect}[2]{\mathbb{E}_{#1}~#2}

\newcommand{\occlusion}{z}

\newcommand{\T}{\mathbf{P}}
\newcommand{\F}{\mathbf{T}}
\icmltitlerunning{Unsupervised Part Representation by Flow Capsules}
\begin{document}

\twocolumn[
\icmltitle{Unsupervised Part Representation by Flow Capsules}

% It is OKAY to include author information, even for blind
% submissions: the style file will automatically remove it for you
% unless you've provided the [accepted] option to the icml2020
% package.

% List of affiliations: The first argument should be a (short)
% identifier you will use later to specify author affiliations
% Academic affiliations should list Department, University, City, Region, Country
% Industry affiliations should list Company, City, Region, Country

% You can specify symbols, otherwise they are numbered in order.
% Ideally, you should not use this facility. Affiliations will be numbered
% in order of appearance and this is the preferred way.

\begin{icmlauthorlist}
\icmlauthor{Sara Sabour}{to,goo}
\icmlauthor{Andrea Tagliasacchi}{to,goo}
\icmlauthor{Soroosh Yazdani}{goo}
\icmlauthor{Geoffrey E. Hinton}{to,goo}
\icmlauthor{David J. Fleet}{to,goo}
\end{icmlauthorlist}

\icmlaffiliation{to}{Department of Computer Science, University of Toronto}
\icmlaffiliation{goo}{Google Research, Brain}

\icmlcorrespondingauthor{Sara Sabour}{sasabour@google.com}

% You may provide any keywords that you
% find helpful for describing your paper; these are used to populate
% the "keywords" metadata in the PDF but will not be shown in the document
\icmlkeywords{Machine Learning}

\vskip 0.3in
]

% this must go after the closing bracket ] following \twocolumn[ ...

% This command actually creates the footnote in the first column
% listing the affiliations and the copyright notice.
% The command takes one argument, which is text to display at the start of the footnote.
% The \icmlEqualContribution command is standard text for equal contribution.
% Remove it (just {}) if you do not need this facility.

\printAffiliationsAndNotice{}  % leave blank if no need to mention equal contribution
%\printAffiliationsAndNotice{\icmlEqualContribution} % otherwise use the standard text.

\begin{abstract}
\vspace{-0.15cm}
Capsule networks aim to parse images into a hierarchy of objects, parts and relations.
While promising, they remain limited by an inability to learn effective low level part descriptions.
To address this issue we propose a way to learn primary capsule encoders that 
detect atomic parts from a single image.
During training we exploit motion as a powerful perceptual cue for part definition, 
with an expressive decoder for part generation within a layered image model with occlusion.
Experiments demonstrate robust part discovery in the presence of multiple objects, cluttered 
backgrounds, and occlusion. The  part decoder infers the underlying shape 
masks, effectively filling in occluded regions of the detected shapes.
We evaluate FlowCapsules on unsupervised part segmentation and unsupervised image classification.
\vspace{-0.2cm}
%
% Capsule networks are designed to parse an image into a hierarchy of objects, parts and relations.
% While promising, they remain limited by an inability to learn effective low level part descriptions.
% To address this issue we propose a novel self-supervised method for learning part descriptors of an image.
% During training, we exploit motion as a powerful perceptual cue for part definition, using an expressive decoder 
% for part generation and layered image formation with occlusion.
% Experiments demonstrate robust part discovery in the presence of multiple objects, cluttered 
% backgrounds, and significant occlusion.
% The resulting part descriptors, a.k.a. part capsules, are decoded into shape masks, filling in occluded 
% pixels, along with relative depth on single images.
% We evaluate our part capsules in both tasks of unsupervised part segmentation and unsupervised image classification.
\end{abstract}

\vspace*{-0.6cm}
\section{Introduction}
\label{sec:intro}
\vspace*{-0.1cm}

% \todo{Mention amodal completion in sec 5.2?  Note on architecture and related encoder-implicit decoder (im-net \cite{imnet}, occnet \cite{OccNet2019})?
% Differences may be dientangling different parts, and part shapes from the pose?}

Humans learn to perceive shapes in terms of  parts and their spatial relationships~\cite{Hoffman2001}.
Studies show that infants form early object perception by dividing visual inputs into units that move rigidly 
and separately~\citep{spelke1990principles}, and they do so in a largely unsupervised way.
Inspired by this and recent work on part discovery, we propose a self-supervised way to learn visual part descriptors for Capsule networks~\cite{Hinton2011tae}.

Capsule networks represent objects in terms of primary part descriptors in a local canonical frame, 
and coordinate transformations between parts and the whole.  As a result of their architecture, 
they are robust to various challenges, including viewpoint changes and adversarial attacks.
Stacked capsule network architectures (SCAE)~\citep{kosiorek2019stacked} have shown promising results 
on a number of simple image datasets. Nevertheless, because they are trained with an image reconstruction 
loss, foreground-background separation and part discovery in cluttered images remain challenging.

\begin{figure}[t]
\vspace*{-0.15cm}
\begin{center}
        \begin{overpic}
            [width=0.9\linewidth]            
            {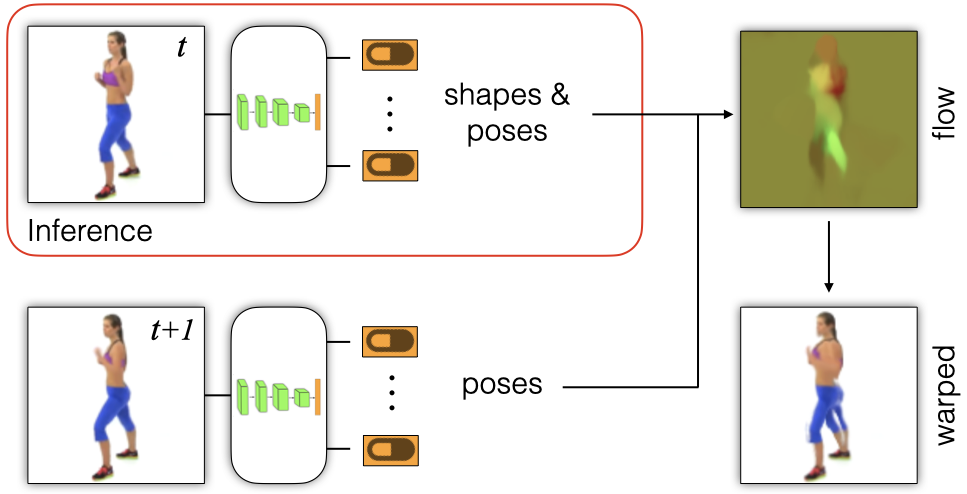}
            % {fig/teaser3.jpeg}
        \end{overpic}
\end{center}
\vspace*{-0.3cm}
\caption{
\textbf{Self-supervised training for learning primary capsules:} 
A single image encoder is trained to decompose the scene into a collection of \textit{primary capsules}.
Learning is accomplished in an unsupervised manner, using flow estimation from capsule shapes and poses as a \textit{proxy} task.
}
\label{fig:teaser}
\vspace*{-0.3cm}
\end{figure}

This paper introduces a way to learn encoders for object parts (aka., primary capsules) 
to address these challenges.
The encoder takes as input a single image (see Fig.\ \ref{fig:teaser}), but 
for training part discovery uses motion-based self-supervision~\cite{Bear-NeurIPS20,Mahendran2018}.
Like the classical literature on perceptual organization and common fate in Gestalt psychology~\citep{spelke1990principles,WagemansPsychBull2012}, we exploit the fact that regions 
of the image that move together often belong together.  
This is a strong perceptual cue that facilitates foreground-background segmentation
and part discovery, and allows one to disentangle texture and other aspects of appearance from shape.

The proposed part encoder (Fig.\ \ref{fig:archTest}) captures the underlying part shapes, their relative poses, 
and their relative depth ordering.
The introduction of depth ordering is particularly useful in order to account for occlusion, 
as it is in classical layered motion models \citep{WangAdelson1994}.
In this way, learning aggregates information about shape over many images,
even though a given part may rarely be visible in its entirety in any single frame.  
In essence, the model prefers simple part-based descriptions, where many variations in appearance
can be explained by a coordinate transform or by occlusion, rather than by changes in shape.

We demonstrate the FlowCapsules approach on several datasets showcasing challenges due to 
texture, occlusions, scale, and instance variation. We compare FlowCapsules to recent 
related work including PSD \citep{xu2019unsupervised} and R-NEM \cite{van2018relational}, 
where part masks and dynamics are learnt using motion.
FlowCapsules provide unsupervised shape segmentation, even in the face of texture 
and backgrounds, outperforming PSD \citep{xu2019unsupervised}.
FlowCapsules also provide a depth ordering to account for occlusion, with the added benefit that
part inference yields shape completion when parts are partially occluded. 

We also report unsupervised classification of images using FlowCapsules part embeddings. 
We compare our results on several datasets with different challenges against SCAE~\citep{kosiorek2019stacked}. 
Experiments show that FlowCapsules consistently outperform SCAE in unsupervised object classification, 
especially on images with textured backgrounds.

\vspace*{-0.1cm}
\section{Related Work}
\vspace*{-0.1cm}

Given the vast literature of part-based visual representations,
we focus here only on the most closely related recent work.

Transforming autoencoders \citep{Hinton2011tae} introduced capsule networks. 
\citet{Sabour2017capsule} revisited the capsule concept and introduced capsule hierarchies 
for object classification, and subsequent work has produced improved routing algorithms \citep{Hinton2018capsule,hahn2019self,ahmed2019star}.
Nevertheless, learning primary capsules from images has remained largely untouched. 
An analogy to text understanding would be a language with a well defined 
grammar and parser, but no good definition or representation of words. 
We introduce a technique for learning primary capsules to address this shortcoming.

Unsupervised capsule learning with an \textit{image reconstruction} loss for part discovery
has been explored by \cite{kosiorek2019stacked} and \cite{Rawlinson2018sparsecaps}.
Several works learn capsule autoencoders for 3D objects from point clouds~\citep{srivastava2019geometric,zhao20193d,sun2020caca}.
But with the exception of capsule models trained with class labels \citep{Hinton2018capsule} 
or segmentation masks \citep{Lalonde2018capsule, Duarte}, previous methods struggle with natural images. 
Object-background discrimination with cluttered, textured scenes is challenging for an image
reconstruction loss. With self-supervised training and visual motion, FlowCapsules achieve
part discovery without ground truth labels or segmentation masks.

Recent approaches to object-centric learning, e.g., MONet \citep{Burgess2019monet}, IODINE~\citep{Greff2019multi}, 
and Slot-attention~\citep{locatello2020object}, focus on learning object representations via image reconstruction.
Beyond the need to reconstruct image backgrounds, they require iterative refinement for
symmetry breaking and forcing scenes into slots. 
In contrast, FlowCapsule learning relies on reconstruction of the flow rather than the image,
and with motion as the primary cue, scenes are decomposed into parts without needing iterative refinement. 
Most recently, \citep{Bear-NeurIPS20, veerapaneni2020entity} extend such networks to incorporate motion,
but still rely on iterative refinement. 
FlowCapsule encodings further disentangle shape and pose, enabling shape completion during partial occlusion.

FlowCapsules currently represent 2D objects, reminiscent of layered models but with a feedforward encoder.
Classical layered models \cite{WangAdelson1994,SpritesCVPR2001} used mixture models and assigned pixels to layers independently, often failing to capture the coherence or compactness of object occupancy.
Some methods use  MRFs to encourage spatial coherence \citep{WeissCVPR1997}.  Others  
enforce coherence via local parametric masks~\citep{Jepson2002}.

% \at{We note that successful detection frameworks like \cite{DPM2010} or \cite{ISM2004} also use object-part models, but these use a probabilistic constellation of parts comprise relatively sparse feature activations that, differently from our method, express no concrete notion of shape.}
% are constructed by models for objects as a probabilistic constellation of parts
% have been proposed with object-part models.
% to deal with category diversity, as well as object flexibility and articulation.
% Early examples include the \cite{DPM2010} % Deformable Parts Models
% and \cite{ISM2004}, % Implicit Shape Models
% both of which were successful in detecting  complex objects in natural images.  Such models learn models for objects as a probabilistic constellation of parts comprising relatively sparse feature activations, but with no concrete notion of shape.

\begin{figure*}[th]
\begin{center}
\begin{overpic}
    [width=0.825\linewidth]
    {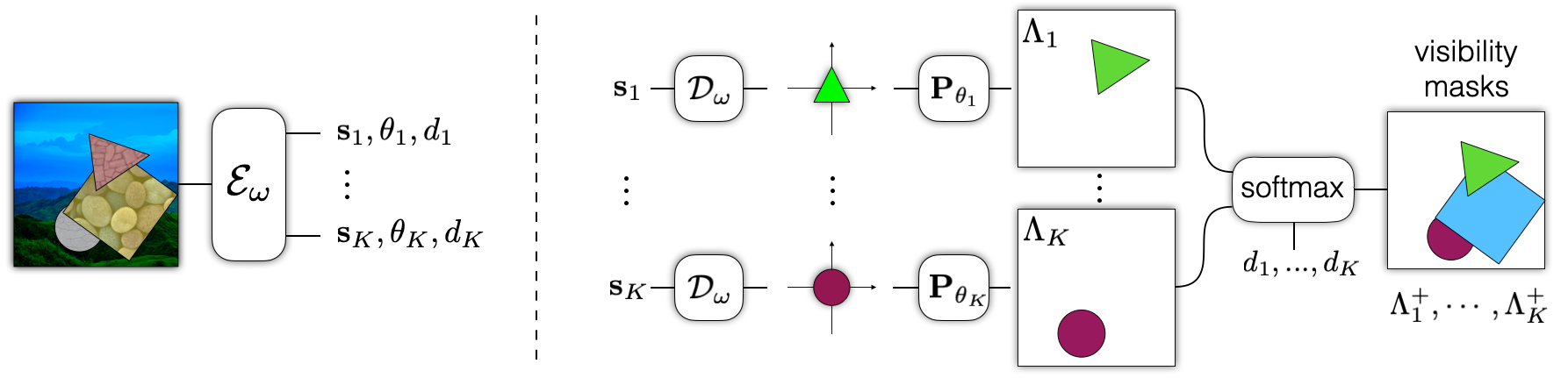}
\end{overpic}
\end{center}
\vspace*{-0.35cm}
\caption{
\textbf{Inference architecture.}
(left) The encoder $\encoder$ parses an image into part capsules, each 
comprising a shape vector $s_k$, a pose $\theta_k$, and a scalar depth value $d_k$. 
(right) The shape decoder $\mathcal{D}_\omega$ is an implicit function.  
It takes as input a shape vector, $s_k$, and a location in canonical 
coordinates and returns the probability that the location is inside the shape.
Shapes are mapped to image coordinates, using $\theta_k$, and layered 
according to the relative depths $d_k$, yielding visibility masks.
}
\label{fig:archTest}
\vspace*{-0.2cm}
\end{figure*}

Visual motion is well-known to be a strong cue for self-supervised learning.
For example, \cite{vijayanarasimhan2017sfm} learn to infer depth, segmentation, and relative 3D motion from 
consecutive frames using self-supervised learning with photometric constraints.
\comment{Given a sequence of frames, SfM-Net \cite{vijayanarasimhan2017sfm} predicts depth, segmentation, camera and 
rigid object motion using motion information.  Their model can be trained with various degrees of supervision,
including 1) self-supervised with familiar re-projection and a photometric loss, 2) supervised with ego-motion
(camera motion), or 3) supervised by depth (e.g., given RGBD data).
SfM-Net extracts meaningful depth estimates and successfully estimates frame-to-frame camera rotations and translations. 
It often successfully segments the moving objects in the scene, even though such supervision is never provided.}
These and related methods use \textit{optical flow} or multiple frames as an input.
FlowCapsules use video frame pairs during training, but the part encoder (see Fig.\ \ref{fig:archTest}),
takes as input a \textit{single} frame. In essence, it learns to decompose images into {\em movable} objects.

S3CNNs \cite{Mahendran2018} takes a similar approach, but does not learn per-part shape encodes or coordinate frames. Rather, they learn to group pixels using patch-wise affine flow, rather than expressing flow in terms of coherent parts and their coordinate frames.
A closely related method is PSD~\citep{xu2019unsupervised}, which 
% , with which we compare in experiments below. PSD
uses optical flow to learn hierarchical part-based models of shape and dynamics in a layered image model.  
It trains a VAE flow encoder and an image encoder to predict the next frame. 
Both PSD and S3CNNs require ground truth flow during training and lack an explicit canonical 
part descriptor like FlowCapsules.

% \TODO{Above needs work. Discuss \cite{Bear-NeurIPS20} and \cite{Greff2019multi} and others (maybe slot attention).}

Our work is also related to generative shape models.
% (e.g., \citet{EslamiNIPS12}).
\citet{HuangMurphyICLR16} learn parts in a layered model with depth order and occlusion.  
Given an image, \textit{variational} inference is used to infer shape and foreground/background separation. 
FlowCapsule encoders, by comparison, are trained as auto-encoders and are therefore easier to learn.
Several recent papers learn generative models that \textit{disentangle} shape and deformation ~\cite{SkafteHauberg2019,deng2021generative}.
FlowCapsules disentangle shape and transformations from canonical to image coordinates. In doing so they
decompose shapes into multiple near-rigid parts with occlusions. 
FlowCapsules thereby disentangle shape at a finer granularity. 
Also, \citet{SkafteHauberg2019} and \citet{deng2021generative} use an  image reconstruction loss, much like SCAE, while FlowCapsules 
only encode shape silhouettes, which simplifies training and the disentangled representation.

%\clearpage

\vspace*{-0.1cm}
\section{Model}
\vspace*{-0.15cm}

Our goal is to learn an encoder that parses images of familiar shapes into parts.
% , depicted in  Fig.\ \ref{fig:archTest}.
To facilitate training, and downstream tasks, we also learn a decoder capable of 
generating segment masks for the parts in the image.
In what follows we first describe the form of the proposed capsule encoder and the mask decoder.
The subsequent section then describes the objective and training procedure.

\begin{figure}[t]
\vspace*{-0.2cm}
\begin{center}
        \begin{overpic}
            [width=0.8\linewidth]
            {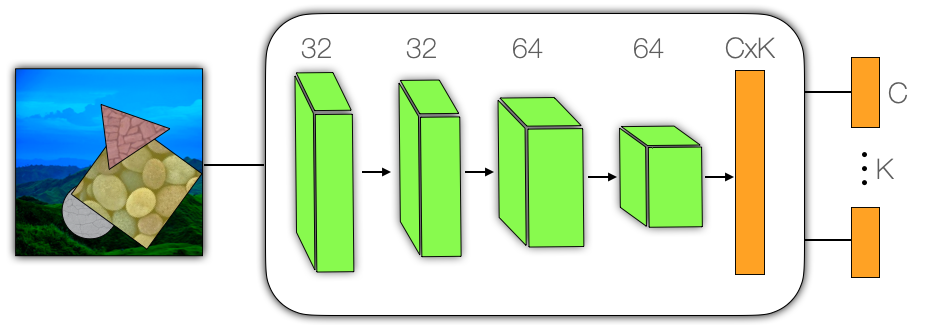}
        \end{overpic}
\end{center}
\vspace*{-0.4cm}
\caption{
\textbf{Encoder architecture.} The encoder comprises convolution layers with ReLU activation,
followed by down-sampling via $2\! \times \!2$ AveragePooling. 
Following the last convolution layer is a \texttt{tanh} fully connected layer,
and a fully connected layer grouped into $K$, $C$-dimensional capsules.
}
\vspace*{-0.3cm}
\label{fig:archEncode}
\end{figure}

\paragraph{Image encoder}
The capsule encoder $\encoder$, with parameters $\params$, encodes a given image  a collection of $K$ primary capsules.
The architecture we propose is depicted in Figure~\ref{fig:archEncode}.
Each capsule, $\capsule_k$, comprises a vector $\shape_k$ that encodes the \textit{shape} of the part,
a \textit{pose} vector $\pose_k$, and a depth scalar $\depth_k$:
\begin{equation}
    \encoder(I) = \{ \capsule_0, \dots, \capsule_k \}, \quad \capsule_k = (\shape_k, \pose_k, \depth_k) ~.
\end{equation}
Capsule shapes are encoded in a canonical coordinate frame.  
The scalar $\depth_k$ specifies relative inverse depth (larger for foreground objects).
The pose vector specifies a mapping from part-centric coordinates ${\bf v}$ to image coordinates $\pixel$
(or scene coordinates more generally), i.e., $\pixel = \T_{\pose_k} {\bf v}$.

As we focus on planar layered models with depth $\depth$, we define $\T_{\pose_k}$ to be a conformal map.  
Accordingly, let $\pose_k {\in} \Real^4$, where $[\pose_k]_{0,1}$ represents the translation,  $[\pose_k]_{2}$ is the rotation angle, and $[\pose_k]_{3}$ is the change in scale. 
More concretely (subscript $k$ is dropped for readability):
\begin{align}
    \T_{\pose} \,  = \,
    \begin{bmatrix}
   \pose_3\cos(\pose_2) & -\pose_3 \sin(\pose_2) & \pose_0\\ 
   \pose_3\sin(\pose_2) & \pose_3 \cos(\pose_2) & \pose_1 \\
   0 & 0 & 1
   \end{bmatrix}
\end{align}
Taken together, the capsule codes are of size $\capsule_k {\in} {\Real}^C $, 
where $\pose_k {\in} \Real^4$, $\depth_k {\in} \Real$, and therefore $\shape_k {\in} \Real^{C-5}$.

\begin{figure*}[t]
\begin{center}
        \begin{overpic}
            [width=0.75 \linewidth]
            {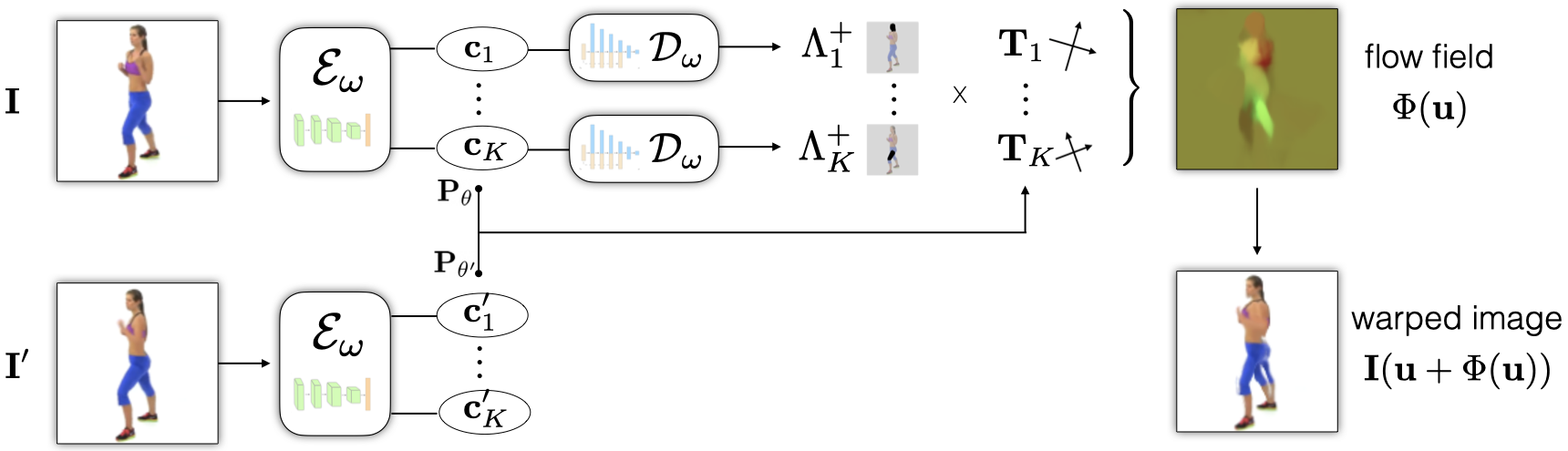}
        \end{overpic}
\end{center}
\vspace*{-0.25cm}
\caption{
\textbf{Self-supervised training} -- 
% An image encoder $\encoder$ decomposes the scene into a collection of $K$ \textit{primary capsules} 
% $\{\mathbf{c}_k\}$, each comprising a shape encoding, pose and depth.
Training uses a \textit{proxy} motion task in which the capsule encoder is applied to a pair of successive 
video frames, providing $K$ primary capsule encodings from each frame.
Visible part masks, $\Lambda^+_k$,  and their corresponding poses, $ {\T}_{\pose} $, determine a flow field 
$\flow$ that is used to warp image $\image$ to predict $\next\image$ in the loss
$\mathcal{L}_\text{render}$ in (\ref{eq:loss_render}).
}
\vspace*{-0.2cm}
\label{fig:archTrain}
\end{figure*}
%We seek to train a single image encoder $\encoder$ capable to decompose the scene into a collection of $K$ \textit{primary capsules} $\{\mathbf{c}_k\}$.
%We perform this task via the \textit{proxy} task of regressing a flow field $\flow$ that is used to warp image $\image$ into $\next\image$.
% 
% The flow field is determined by the parts visibility masks  $\Lambda^+_k$ in (\ref{eqn:lambdaplus}), and the part coordinate transforms $\F_k $ in (\ref{eq:partTransform}).
% The rendering loss for training (\ref{eq:loss_render}) is the residual error between the warped version of $\image$ and the next frame $\next\image$.
% 

\paragraph{Mask decoder}
A mask decoder facilitates self-supervised learning of the encoder, as well as downstream segmentation tasks.
It allows one to visualize the part and connect it to observations in the image.
As depicted in \Figure{archTest}, the mask decoder $\masknet$ generates an object silhouette~(or mask) 
in canonical coordinates, which is then mapped to image coordinates, incorporating occlusion and visibility.

\begin{figure}[t]
\vspace*{-0.2cm}
\begin{center}
        \begin{overpic}
            [width=0.7\linewidth]
            {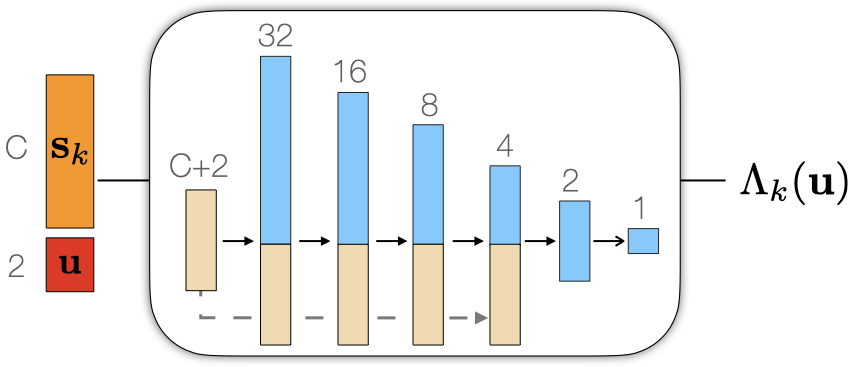}
        \end{overpic}
\end{center}
\vspace*{-0.35cm}
\caption{
\textbf{Decoder architecture.}  A neural implicit function \citep{imnet} is used to represent part masks.
An MLP with SELU activations \cite{klambauer2017self} takes as input a shape vector $s$ and a pixel position ${\bf u}$. 
Applied to a pixel grid, it produces a logit grid for the mask.
}
\vspace*{-0.25cm}
\label{fig:archDecode}
\end{figure}

In more detail, the mask decoder represents the shape of the part in its canonical 
coordinate frame, $ \masknet({\bf v}; \shape_k) $.  
Our current decoder architecture is depicted in \Figure{archDecode}.
This is then mapped into image coordinates according to the pose vector~$\pose_k$, 
yielding the shape mask~$\Lambda_k$ in the image frame:
\begin{align}
    \Lambda_k (\pixel) &~=~ \masknet( \pixel\T^{-1}_{\pose_k} ; \shape_k)  ~,
\end{align}
where the map $\T_{\pose_k}$ has parameters $\pose_k$.
We also note that $\Lambda_{k}$ is a \textit{function} of spatial position and a latent code~\cite{imnet,OccNet2019}, but unlike previous work, our encoder disentangles individual part shapes and their poses with respect to canonical coordinates.
% \todo{As such, this technique is similar to other implicit functions (imnet,occnet), but here we have canonicalized pose and diisentngled parts.
% ie note on architecture and related encoder-implicit decoder (im-net \cite{imnet}, occnet \cite{OccNet2019})?}

%\Lambda^+_k = \text{Softmax}_k(\depth_k \Lambda_k) ~.
    
{\bf Occlusion:} 
With opaque objects, parts will not always will be visible in their entirety.
To account for occlusion, part masks are layered according to their depth order, 
thereby determining the visible portion of each part in a given image.
To ensure differentiable image formation, enabling gradient-based learning, we treat the scalar $\depth_k$ as a logit, and apply a softmax across the logits (depths) at every pixel to generate a visibility mask for each part~\cite{gadelha2019shape}; see~Fig.\ \ref{fig:archTest}.
The visible portion of the $k$-th part is therefore given by 
\begin{equation}
    \Lambda^+_k(\pixel) = \frac{e^{d_k\Lambda_k(\pixel)}}{\sum_{k'} e^{d_{k'} \Lambda_{k'}(\pixel)}}
    \label{eqn:lambdaplus}
\end{equation}
As the gap between the largest $\depth_k$ and other values grows, the softmax approaches the argmax,
which of course would be ideal for opaque layers.

A typical auto-encoder might reconstruct the image in terms of these masks, to formulate an image reconstruction loss.
The problem with such an approach is that the encoder would also need to encode other properties of the images, 
such as texture, lighting and the background, with pixel level accuracy.
To avoid this problem, here we aim only to learn an encoder for the part shapes, positions and depth layering.
To this end we consider a form of self-supervised learning that  relies on primarily on 
motion (optical flow) between consecutive frames in video.
The use of flow provides a strong image cue for the segmentation of parts, {\it without} the need to 
model texture, lighting and other fine-grained properties tied to appearance.

\section{Self-Supervised Learning}
\vspace*{-0.1cm}

Training the capsule encoder exploits motion as a visual cue for separating objects 
and their parts from the immediate background.
To that end, we assume that the training data comprises pairs of adjacent video frames.
Given an image pair, the encoder provides an \textit{ordered} set of capsules for each of the two images.
The poses from corresponding capsules and their visibility masks then determine an optical flow field that is 
used to warp one frame to predict the other. This allows use of brightness constancy and other common objectives 
in optical flow estimation to specify a self-supervised training loss.

In more detail, let the two images of a training pair be denoted $\image$ and $\next\image$.
As shown in \Figure{archTrain}, the capsule encoder extracts an ordered set of capsules from each image.
The part capsules are denoted $\capsule_k {=} (\shape_k, \pose_k, \depth_k)$ and $\capsule_k^\prime {=} (\shape_k^\prime, \pose_k^\prime, \depth_k^\prime)$, for $k \in \{ 1 , ... , K \}$.
From corresponding part capsules we then compute the predicted optical flow $\flow$ from the capsule poses,
yielding a mapping $\F_k$ from one image to the next,
\begin{eqnarray}
\F_k ~=~ ~\T_{\next\pose_k} \circ (\T_{\pose_k})^{-1}   ~.
\label{eq:partTransform} 
\end{eqnarray}
This transform maps image locations in $\image$ to the canonical coordinate 
frame of part $k$, and then into the next frame~$\next\image$.
When combined with the layered visibility masks, this provides the flow field:
\begin{eqnarray}
\flow(\pixel  \, |\, \encoder(\image),\encoder(\next\image) ) ~=~ \sum_{k=1}^{K}  ~
\underbrace{\Lambda^+_k(\pixel)}_{\text{visibility}} \, % _{\pixel \in \capsule_k?}
\underbrace{\left [\F_k(\pixel)-\pixel\right]}_\text{~flow of $k$-th capsule} 
\label{eq:flow_basic}
\end{eqnarray}
where $ \pixel \in [-1,1]^2 $ denotes 2D normalized image coordinates.
Note that the use of $[\F_k(\pixel)-\pixel]$ in \eq{flow_basic} ensures that the 
generation of an \textit{identity} flow is the easiest prediction for the network 
$\F_k(\pixel)$ to make (like a residual connection).

% \comment{Given a normalized coordinate $\pixel$ and (concatenated to) a shape latent code $\shape_k$, it determines whether $\pixel$ belongs to the k-th capsule.
% It does so by first writing the (query) pixel $\pixel$ in the \textit{local} coordinate frame of the k-th capsule -- this is only possible because $\masknet$ is a \textit{function}.
% Also note we slightly abuse notation and with $\pose \cdot \pixel$ we imply the application of the affine transformation implied by $\pose$ to $\pixel$.}
% % ~(i.e.~a differentiable image sampler).

% \comment{Given the ground truth flow between the training image pair, denoted~$\flowgt$, 
% i.e., mapping pixels from $\image$ to pixels in image $\next\image$ (see~\Figure{arch}),
% the L2 flow loss over image coordinates $ \pixel \in [0,1]^2 $, is given by 
% \begin{equation}
% \mathcal{L}_\text{flow} = 
% \expect{\pixel \sim [0,1]^2}
% {\|\,\flow(\pixel\, |\,\encoder(\image),\, \encoder(\next\image)) - \flowgt (\pixel)\, \|^2_2 }  
% ~.
% \label{eq:flowloss}
% \end{equation}
% Alternatively, or in combination, one can also consider the L2 brightness constancy loss:}

Given the estimated flow between a given training pair, we warp the pixels of $\image$ according 
to $\flow$, providing a prediction for $\next\image$. 
Then we optimize an L2 brightness constancy loss on the residual errors between
our warped version of the first frame and the second frame,
\begin{equation}
\mathcal{L}_\text{render} \,=\, \expect{\pixel \sim [0,1]^2}
{\| \,  \image(\pixel + \flow(\pixel)) - \next\image(\pixel) \, \|^2_2} \, ~,
\label{eq:loss_render}
\end{equation}
where we have abbreviated $\flow(\pixel\, |\, \encoder(\image), \encoder(\next\image))$ by $\flow(\pixel)$
for notational simplicity.
% --- OLD VERSION
% \begin{align}
% \mathcal{L}_\text{render} = & \expect{\pixel \sim [0,1]^2}  \nonumber \\
% &{\| \,  \image(\, \pixel +  \flow(\pixel\, |\, \encoder(\image), \encoder(\next\image)) \, ) - \next\image(\pixel) \, \|^2_2} ~.
% \label{eq:loss_render}
% \end{align}

We also exploit two simple but effective regularizers on flow and the canonical shape representation. 
They are useful as we do not make use of ground truth segmentation masks or flow fields during training.
The first regularizer, $\mathcal{L}_\text{smooth}$, is a smoothness term often used in 
optical flow estimation \cite{DerpanisECCV2016} to enhance gradient 
propagation through larger movements and regions with negligible brightness variation; i.e.,
\begin{align}
\mathcal{L}_\text{smooth} \,=\, \left\| \frac{\partial \flow}{\partial u_x}, \frac{\partial \flow}{\partial u_y} \right\|^2_2 ~.
\label{eq:loss_smooth}
\end{align}

The second regularizer encourages part shapes to be centered at the origin in the canonical coordinate frame; i.e.,
\begin{equation}
\mathcal{L}_\text{center} = 
\tfrac{1}{K} \sum_{k=1}^{K} \frac{\sum_{\bf v} \| {\bf v} \, \Lambda_k({\bf v})  \|^2_2 }{\sum_{\bf v'} \Lambda_k({\bf v'})} 
% &\|\frac{\sum_u(u_x-.5)\Lambda_k(u) }{\sum \Lambda_k(u)}, \frac{\sum_u (u_y-.5)\Lambda_k(u)}{\sum \Lambda_k(u)}\|^2_2
\label{eq:loss_center}
\end{equation}
Keeping parts centered at $(0,0)$ improves the inference  of rotations. 
For example, a part located far from the origin can easily be projected outside the image
during training.
Keeping it near the center tends to produce a smoother loss function.
% \footnote{This numerical issue can also be solved by training with a larger grid (padded around the original image).} 
% 
The final loss is a weighted sum of the render loss and the two regularizers.

\comment{, 
and where the subscript~$\omega$ indicates modules that possess trainable network parameters.}
\comment{Hence, the dataset we employ for training $\encoder$ consists of pairs of images $\image$ and~$\next\image$, optionally, the flow~$\flow$ mapping $\image$ into $\next\image$, as well as the identity of the objects within each image.} 
\comment{
In the expressions above, the \textit{shared} encoder $\encoder$ decomposes each image as a collection of $K$ ``primary capsules''.
Each capsule, $\capsule_k$, is characterized by shape vector $\shape_k$ that expresses 
its geometry~(i.e.~silhouette), and a \textit{pose} vector $\pose_k$ that capture the part 
transformation into the scene.  In our 2D image model, the pose includes a translation
to image coordinates and relative depth information: 
% ~(e.g.~translation in a 2D image formation process):
% 
\begin{equation}
    \encoder(\image) = \{ \capsule_0, \dots, \capsule_k \}, \quad \capsule_k = (\shape_k, \pose_k)
\end{equation}
\david{Abuse of notation for $\theta$ may be problematic below?}
}

\comment{
\paragraph{Occlusion modeling}
Occlusion needs to be handed carefully in any optical flow model.
We address this by employing a \textit{soft-occlusion} model developed in the context of differentiable rendering~\todo{[?]}.
By extracting an ``occlusion'' scalar $o_k$ from the latent pose $\pose_k$, we can then write:
\begin{align}
\flow(\pixel) &= \sum_k  
{\color{blue}\tilde{\Lambda}_k(\pixel)}
{\left [\T_k(\pixel)-\pixel\right]} \quad
\begin{cases}
\tilde{\Lambda}_k(\pixel) &= \text{softmin}_k(\delta \occlusion_k \Lambda_k(\pixel))
\\
\T_k(\pixel) &= \next\pose_k \cdot (\pose_k)^{-1}
\end{cases}
\label{eq:flow_basic}
\end{align}
where we use a temperature of $\delta{=}100$ for the softmin operator.
}
% We then adjust the masks based on their order to carve out the occluded parts.
% We use a Softmax over the parts to keep the mask value only if no higher $o_k$ is present at that pixel. 
% \begin{equation}
% \end{equation}
% \TODO{Note the last entry in the pose vector $o_k{=}\mathbf{p}_k[-1]$ is the ``depth order'' $o_k$ which will be used to handle occlusions.}

%\clearpage

\section{Experiments}
\vspace*{-0.1cm}

We evaluate FlowCapsules on images with different dynamics, shapes, backgrounds and textures. 
We consider scenes with multiple occluding geometrical shapes (Geo), Geo with textured shapes
and ImageNet backgrounds (Geo$^+$), and on images of people (Exercise).

\vspace*{-0.1cm}
\paragraph{Geo}
For this synthetic dataset, we use the same code and setup as \cite{xu2019unsupervised},
generating 100k images for training, 1k for validation, and 10k for testing.
Images have different background colors, with geometrical shapes (circle, triangle, square) 
of various colors, scales and positions. Objects in Geo undergo translation from frame to frame.

\vspace*{-0.1cm}
\paragraph{Geo$^+$}
% We further analyze robustness of FlowCapsules to texture and background by augmenting the Geo dataset into 
In this variant, we incorporates  natural image backgrounds (random images from ImageNet~\cite{deng2009imagenet}),
and textured foreground shapes. Textures are random samples from the Brodatz dataset~\cite{picard1993real}.

\vspace*{-0.1cm}
\paragraph{Exercise}
This dataset contains natural images of trainers demonstrating exercises, including 
articulated and out of plane motion (used by \cite{xu2019unsupervised}).
% Ideally, FlowCapsules should automatically learn parts that correspond to rigid body segments between two joints. 
The Exercise dataset has 49356 pairs of images for training, extracted from 20 exercise demo videos. 
The test set has 30 images, for which \citet{xu2019unsupervised} provided ground truth segmentation masks.

\vspace*{-0.1cm}
\paragraph{Experimental setup} 
Models are trained using the Adam optimizer \cite{Kingma2014adam} with a fixed learning rate of $1e{-}4$ for 150 epochs. 
We use $C{=}32$ and $K{=}8$ for Geo models and $C{=}16$ and $K{=}16$ for Exercise model.
Regularization constants for $\mathcal{L}_\text{center}$ and $\mathcal{L}_\text{smooth}$
are $1e{-}2$ and $1e{-}4$.
To calculate the intersection-over-union (IoU) performance measure on visibility masks, 
we normalize and then threshold the masks at $0.5$ to get a binary $(0,1)$ mask.

\begin{figure}[t]
    \centering
    \includegraphics[width=\linewidth]{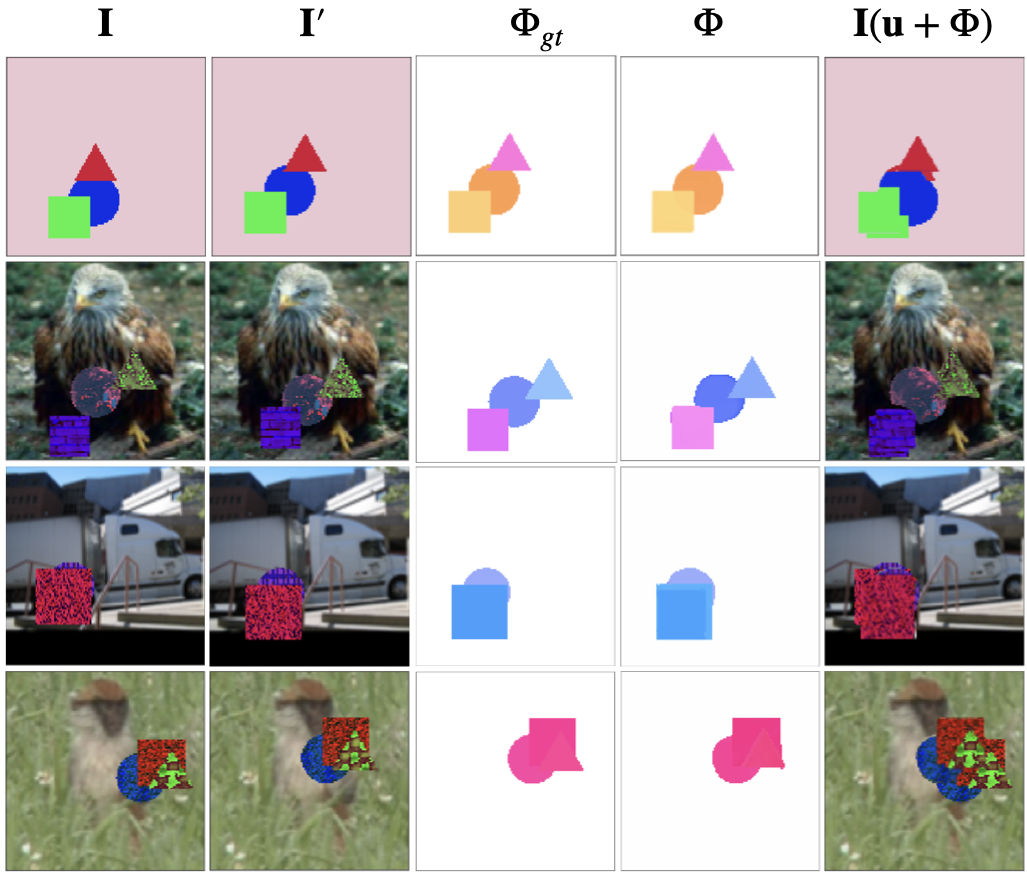}
     \vspace*{-0.5cm}
    \caption{Estimated flows and predicted next frames on training data from \texttt{Geo}~(first row) and \texttt{Geo}$^+$~(rows 2-- 4).
    }
   
    \vspace*{-0.15cm}
    \label{fig:geo_flow}
\end{figure}

\subsection{Estimated Part Motion (\Figure{geo_flow} and \Figure{ex_flow})}
\vspace*{-0.1cm}

To verify that the model estimates flow effectively in an unsupervised manner we first inspect 
the quality of the flow inferred by FlowCapsules after training on each dataset.

\Figure{geo_flow} shows estimated flow $\flow$ alongside the ground truth $\flow_{gt}$  for {\em training} 
image pairs from Geo and Geo$^+$.  The flow is accurate for both datasets.
Comparing the warped version of the first frame $I$ (last column) with the other frame $I'$ 
(second column), one can appreciate some of the challenges in unsupervised flow estimation. 
Because our prediction of~$\next\image$ using~$\flow$ does not account for unoccluded pixels,  $\mathcal{L}_\text{render}$ is not expected to reach $0$. 
We note that while the model uses conformal transformations from frame to frame, these datasets only have
translation; for these data our model correctly estimates zero rotation and unit scale.

\begin{figure}[t]
    \centering
    \includegraphics[width=.975\linewidth]{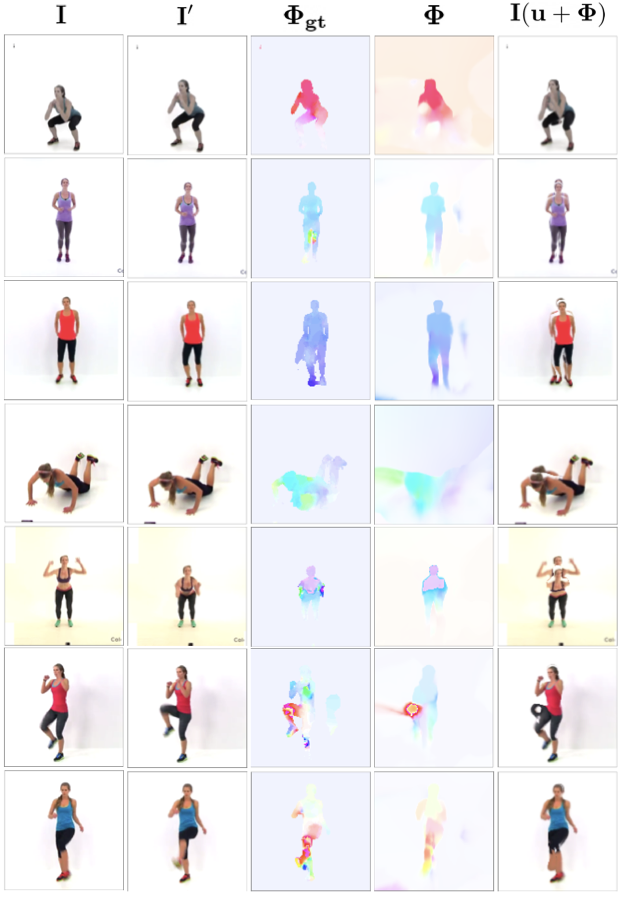}
     \vspace*{-0.1cm}
    \caption{
    Estimated flows and predicted frames on randomly selected images from the \texttt{Exercise} \textit{validation} set. 
    Approximating articulated motion with conformal maps yields reasonable flow fields. 
    The goal is not the best possible flow estimation, but rather, 
    as long as different parts have different flow estimates, our encoder is be able 
    to learn the correct part decomposition.
    }
    \vspace*{-0.3cm}
    \label{fig:ex_flow}
\end{figure}

\comment{
\begin{table}[t]
    \centering
    \begin{tabular}{ccccc}
    $\image$ & $\next\image$ & $\flowgt$ & $\flow$ & $\image(\pixel +  \flow)$ \\
        \includegraphics[width=.15\linewidth]{fig/flow/s_01.png} & 
        \includegraphics[width=.15\linewidth]{fig/flow/s_00.png} &  
        \includegraphics[width=.15\linewidth]{fig/flow/s_0f.png} & 
        \includegraphics[width=.15\linewidth]{fig/flow/s_0pf.png} & 
        \includegraphics[width=.15\linewidth]{fig/flow/s_0p.png}    \\
        \includegraphics[width=.15\linewidth]{fig/flow/bg_00.png} &  \includegraphics[width=.15\linewidth]{fig/flow/bg_01.png} & \includegraphics[width=.15\linewidth]{fig/flow/bg_0f.png} & \includegraphics[width=.15\linewidth]{fig/flow/bg_0pf.png} & \includegraphics[width=.15\linewidth]{fig/flow/bg_0p.png}    \\ 
      \includegraphics[width=.15\linewidth]{fig/flow/bg_30.png} &  \includegraphics[width=.15\linewidth]{fig/flow/bg_31.png} & \includegraphics[width=.15\linewidth]{fig/flow/bg_3pf.png} & \includegraphics[width=.15\linewidth]{fig/flow/bg_3f.png} & \includegraphics[width=.15\linewidth]{fig/flow/bg_3p.png}    \\
      \includegraphics[width=.15\linewidth]{fig/flow/bg_41.png} &  \includegraphics[width=.15\linewidth]{fig/flow/bg_40.png} & \includegraphics[width=.15\linewidth]{fig/flow/bg_4pf.png} & \includegraphics[width=.15\linewidth]{fig/flow/bg_4f.png} & \includegraphics[width=.15\linewidth]{fig/flow/bg_4p.png}    \\
     
    \end{tabular}
    \captionof{figure}{Estimated flows and predicted next frames on \texttt{Geo}~(first row) and \texttt{Geo}$^+$~(remaining rows) \textit{training} set.}
    \label{fig:geo_flow}
\end{table}

\begin{table}[h]
    \centering
    \begin{tabular}{ccccc}
    $\image$ & $\next\image$ & $\flowgt$ & $\flow$ & $\image(\pixel +  \flow)$ \\
        \includegraphics[width=.15\linewidth]{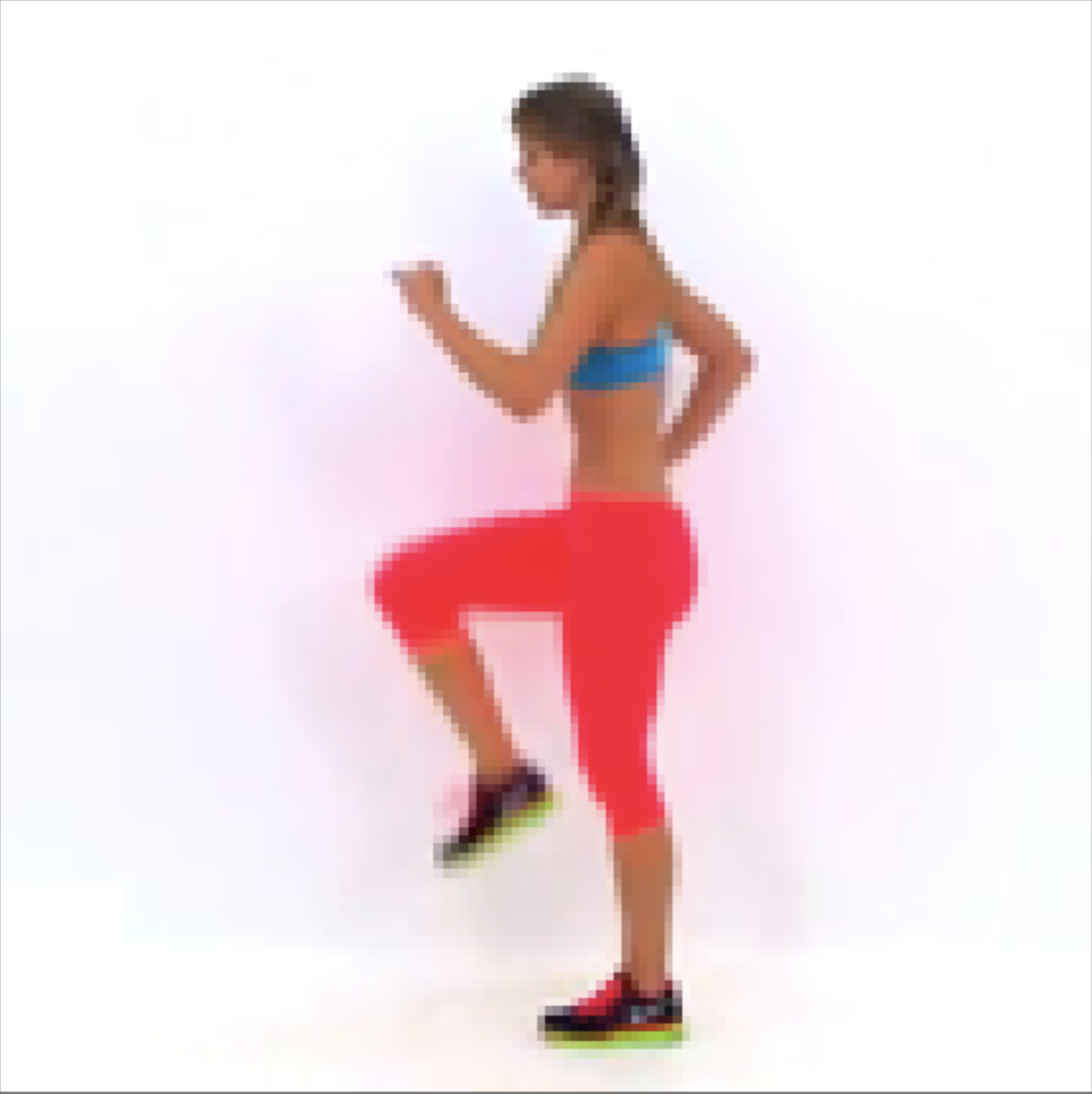} &  \includegraphics[width=.15\linewidth]{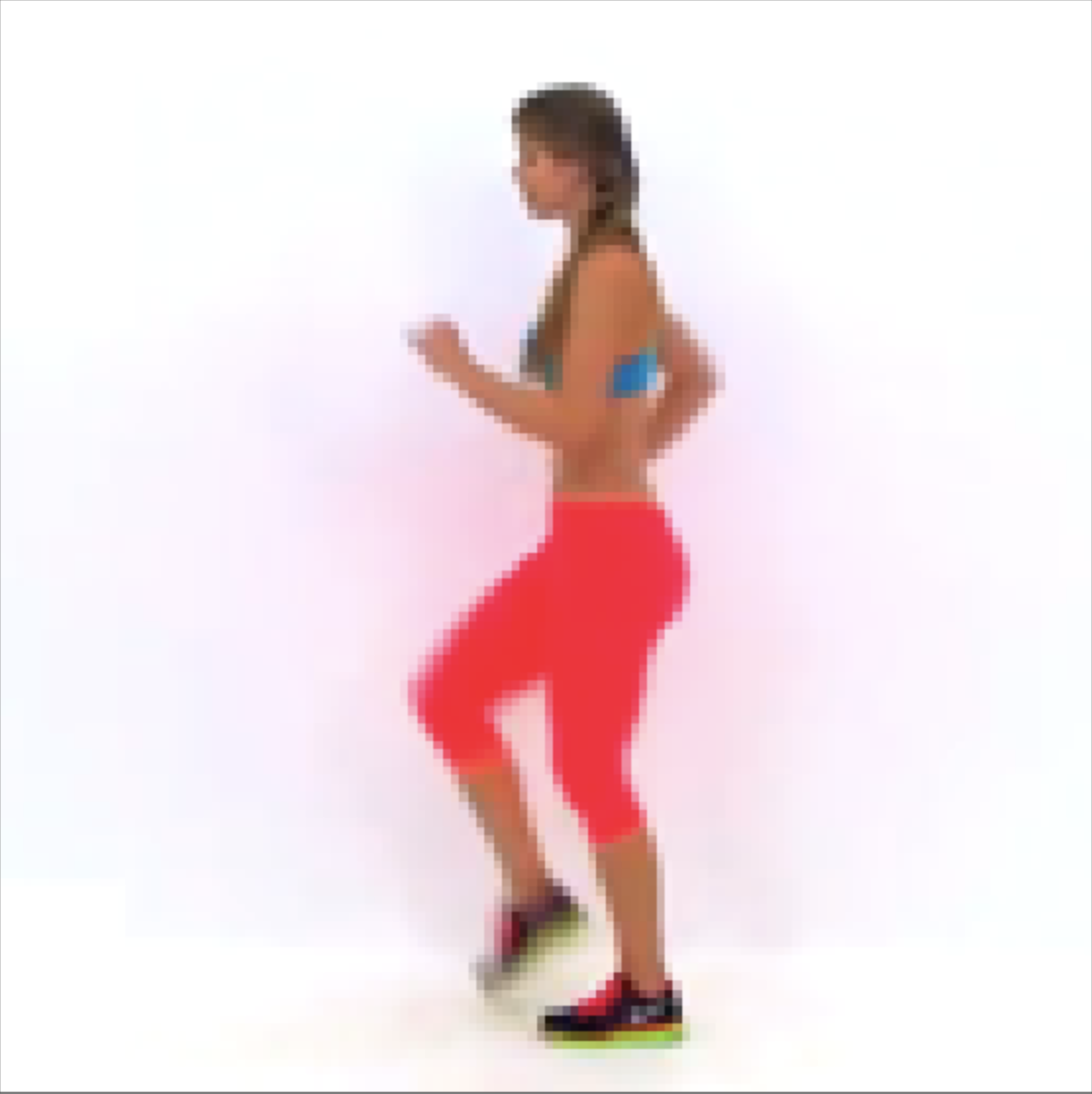} & \includegraphics[width=.15\linewidth]{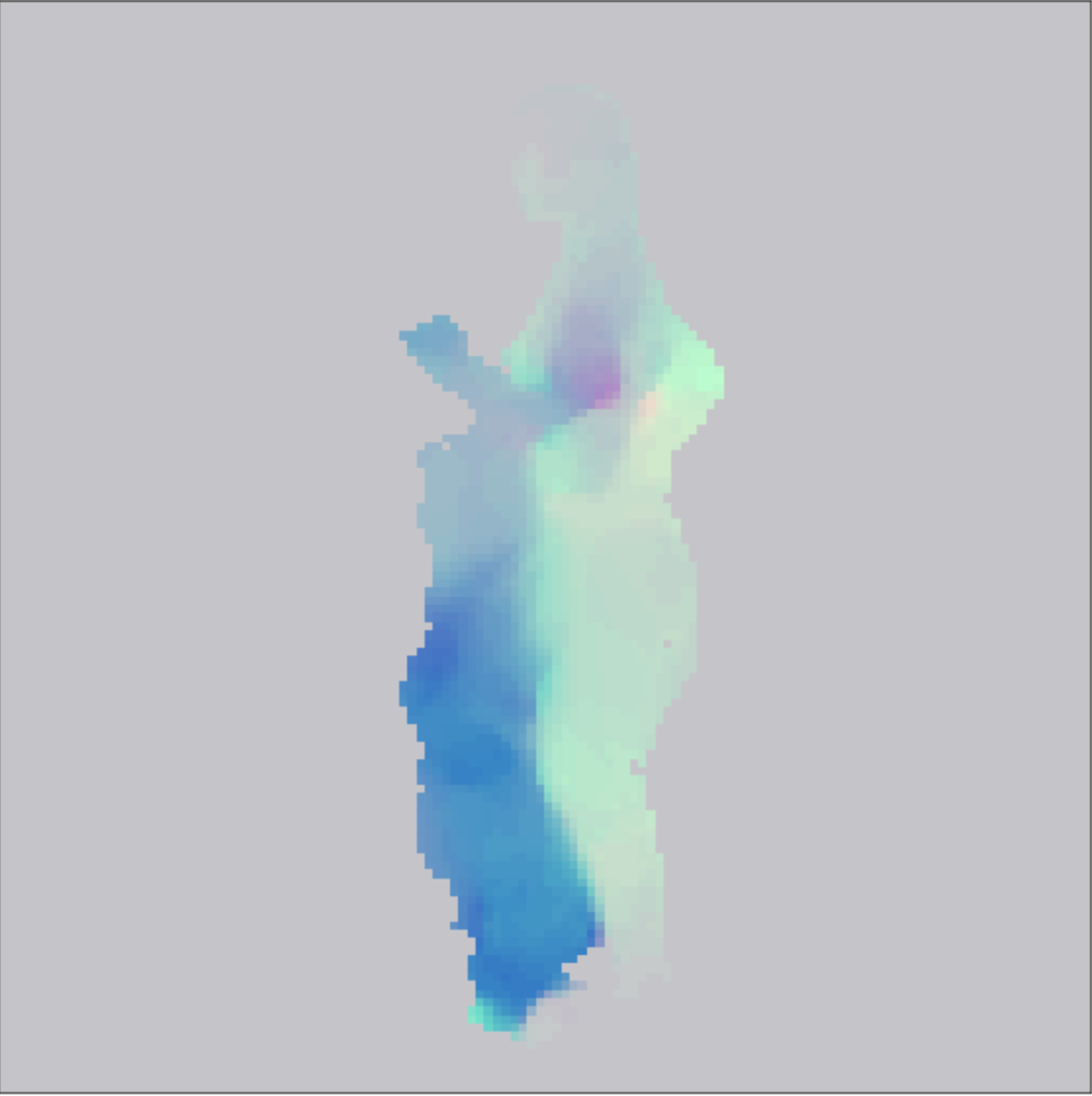} & \includegraphics[width=.15\linewidth]{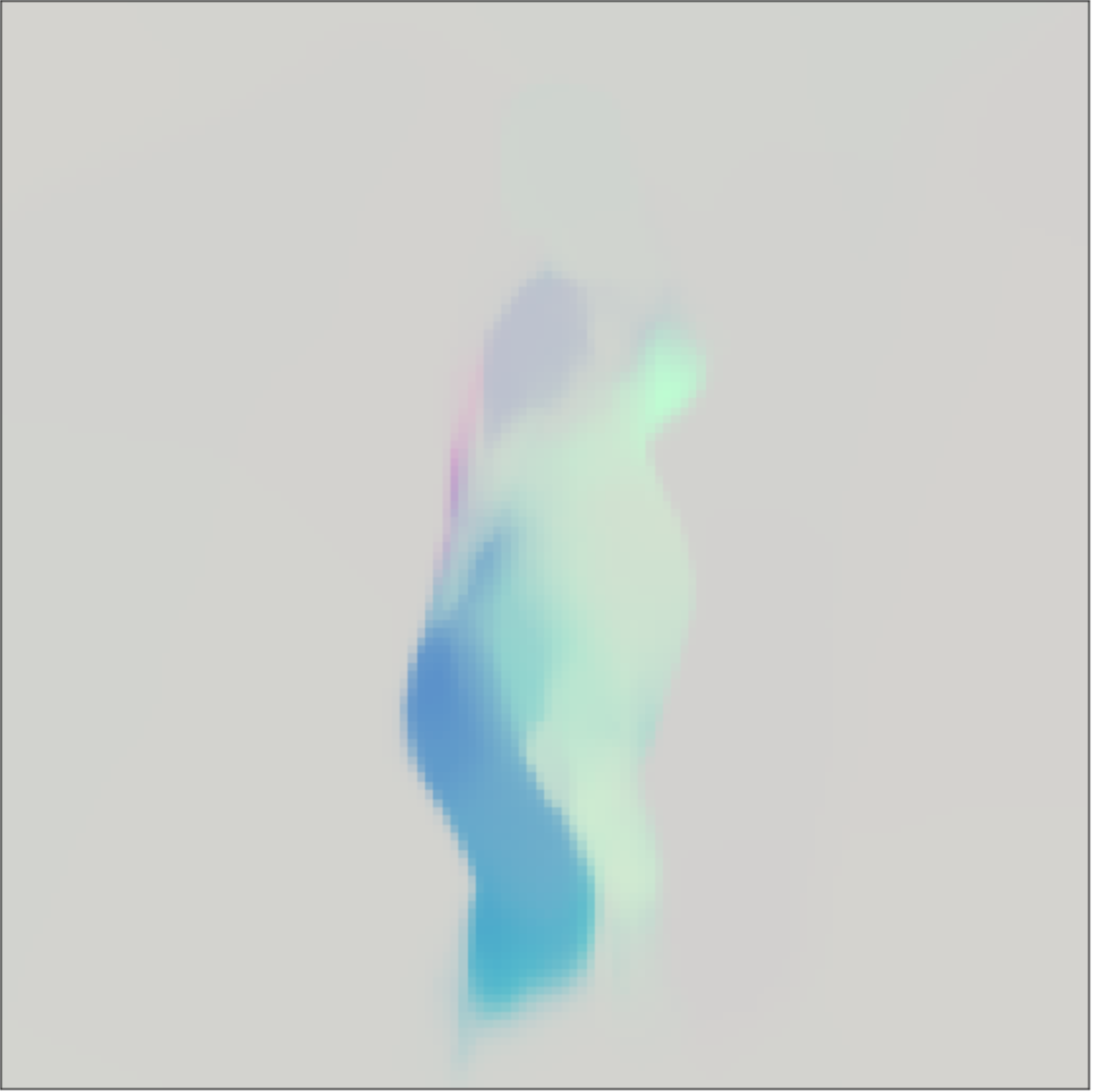} & \includegraphics[width=.15\linewidth]{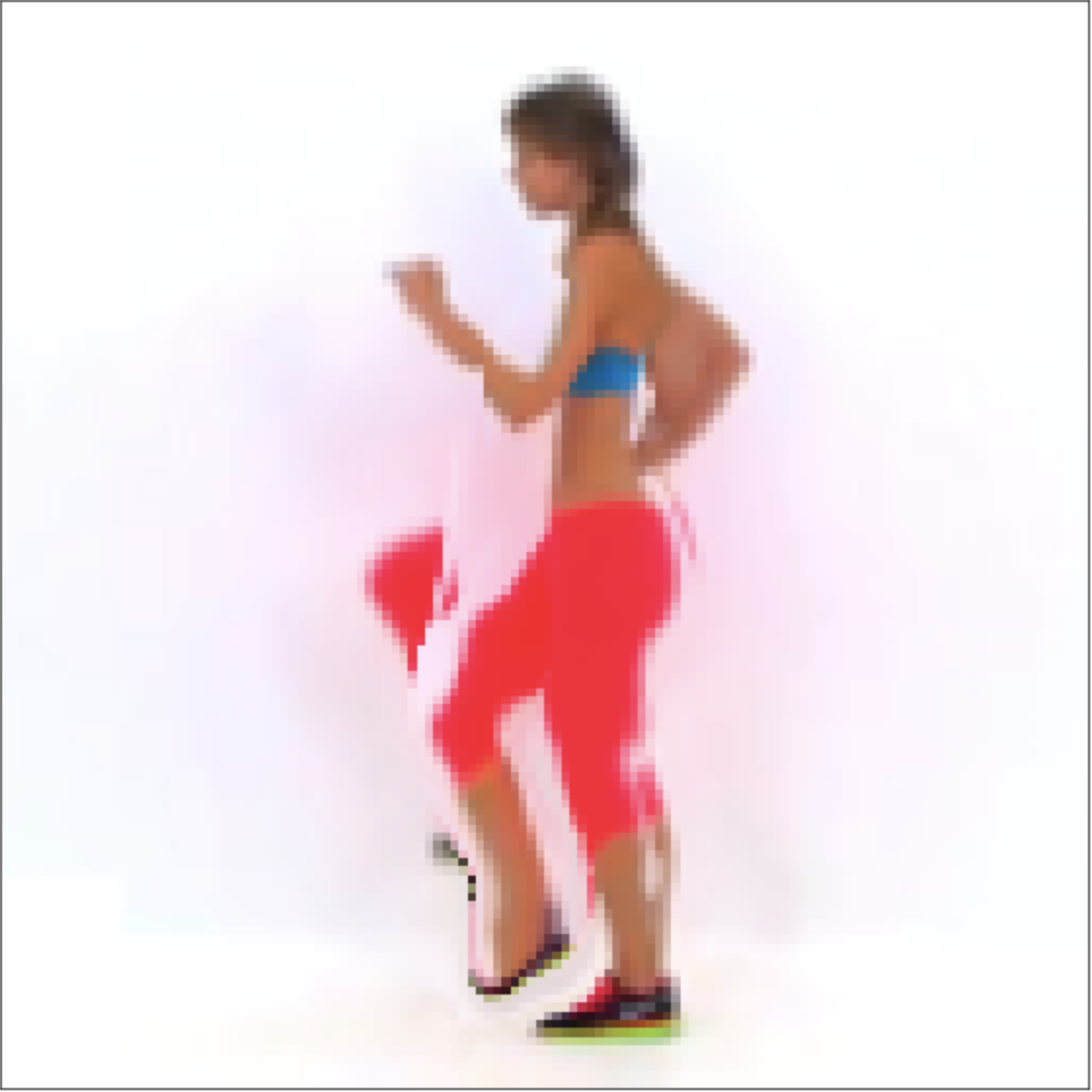}    \\
         \includegraphics[width=.15\linewidth]{fig/flow/1_t0.png} &  \includegraphics[width=.15\linewidth]{fig/flow/1_t1.png} & \includegraphics[width=.15\linewidth]{fig/flow/1_f.png} & \includegraphics[width=.15\linewidth]{fig/flow/1_pf.png} & \includegraphics[width=.15\linewidth]{fig/flow/1_p.png}    \\
         \includegraphics[width=.15\linewidth]{fig/flow/2_t0.png} &  \includegraphics[width=.15\linewidth]{fig/flow/2_t1.png} & \includegraphics[width=.15\linewidth]{fig/flow/2_f.png} & \includegraphics[width=.15\linewidth]{fig/flow/2_pf.png} & \includegraphics[width=.15\linewidth]{fig/flow/2_p.png}    \\
         \includegraphics[width=.15\linewidth]{fig/flow/0_i0.png} &  \includegraphics[width=.15\linewidth]{fig/flow/0_i1.png} & \includegraphics[width=.15\linewidth]{fig/flow/0_f.png} & \includegraphics[width=.15\linewidth]{fig/flow/0_pf.png} & \includegraphics[width=.15\linewidth]{fig/flow/0_p.png}    \\
         \includegraphics[width=.15\linewidth]{fig/flow/3_t0.png} &  \includegraphics[width=.15\linewidth]{fig/flow/3_t1.png} & \includegraphics[width=.15\linewidth]{fig/flow/3_f.png} & \includegraphics[width=.15\linewidth]{fig/flow/3_pf.png} & \includegraphics[width=.15\linewidth]{fig/flow/3_p.png}    \\
         \includegraphics[width=.15\linewidth]{fig/flow/4_t0.png} &  \includegraphics[width=.15\linewidth]{fig/flow/4_t1.png} & \includegraphics[width=.15\linewidth]{fig/flow/4_f.png} & \includegraphics[width=.15\linewidth]{fig/flow/4_pf.png} & \includegraphics[width=.15\linewidth]{fig/flow/4_p.png}    \\
         \includegraphics[width=.15\linewidth]{fig/flow/5_t0.png} &  \includegraphics[width=.15\linewidth]{fig/flow/5_t1.png} & \includegraphics[width=.15\linewidth]{fig/flow/5_f.png} & \includegraphics[width=.15\linewidth]{fig/flow/5_pf.png} & \includegraphics[width=.15\linewidth]{fig/flow/5_p.png}    \\
    \end{tabular}
    \vspace*{-0.5cm}
    \captionof{figure}{Estimated flows and predicted next frames on the \texttt{Exercise} \textit{training} set.}
    \vspace*{-0.15cm}
    \label{fig:ex_flow}
\end{table}
}

\Figure{ex_flow} shows examples of  model flow estimates for the Exercise dataset.
The true flow here reflects the articulated motion of the people, and it is notable
that the parts here are much smaller than those in Geo/Geo$^+$.
Although the estimated flows are somewhat blurred, they still capture different movements  
of different parts reasonably well, even though the model is limited to conformal deformations 
from one frame to the next.

\vspace*{-0.1cm}
\subsection{Unsupervised Part Segmentation}
\vspace*{-0.1cm}

One effective way to evaluate FlowCapsules is to see how well it learns to 
decompose a single image into its \textit{movable} parts.  
We view this as an unsupervised part segmentation task
and we note that, while trained on image pairs, inference is performed on a \textit{single} 
test image, yielding part shapes and a coordinate transform for each part.
Conversely, methods relying on optical flow only generate masks for parts \textit{in motion}, 
as these models effectively build masks by \textit{segmenting} the flow \cite{xu2019unsupervised}.

\begin{figure}[ht]
    \centering
    \includegraphics[width=\linewidth]{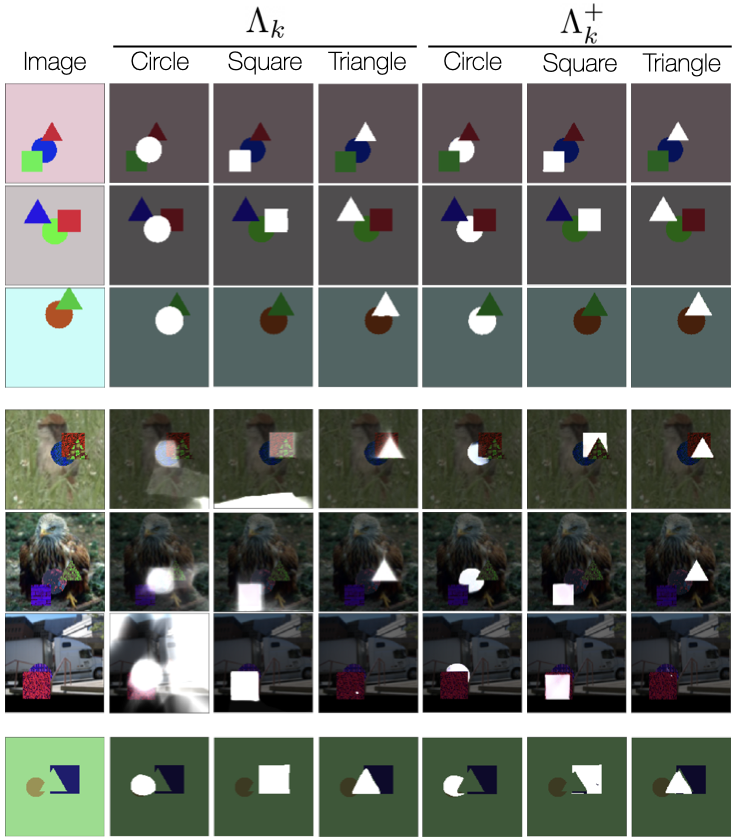}
    \vspace*{-0.4cm}
    \caption{Inferred FlowCapsule shapes and corresponding visibility masks on Geo (rows 1--3), and Geo$^+$ (rows 4--6).
    The third row for each dataset shows an instance with only two objects present, so one mask is empty.
    The last row shows an interesting case in which the 
    triangle is detected by the encoder even though it shares the color of the background, reminiscent of subjective contours \cite{Kanizsa1976}.
    }
    \label{fig:geo_masks}
\end{figure}

\begin{figure*}[ht]
    \centering
    \includegraphics[width=0.99\linewidth]{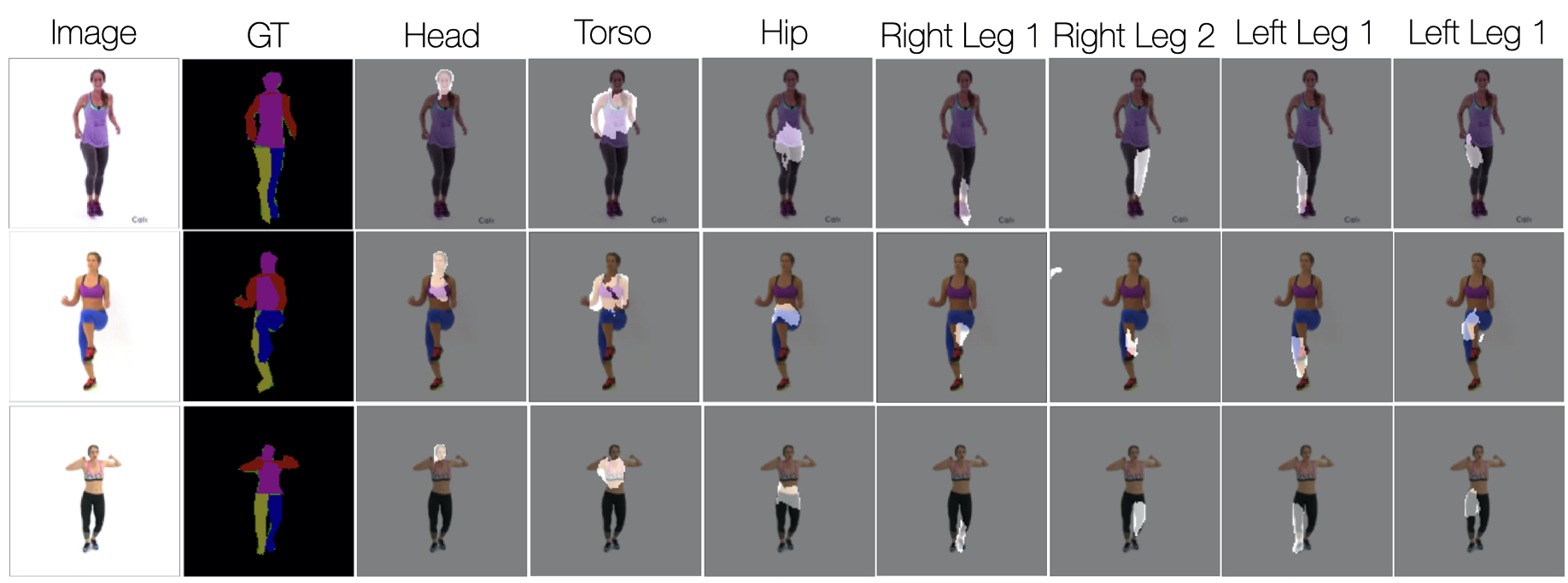}
    \vspace*{-0.1cm}
    \caption{The ground truth segment masks along with sample FlowCapsule masks $\Lambda_k^+$ on Exercise test data.}
    \label{fig:ex_masks}
\end{figure*}

\comment{
\begin{table*}[ht]
    \centering
    \begin{tabular}{ccccccc}
    $\image$ & \multicolumn{3}{c}{ $ \Lambda_k$} &  \multicolumn{3}{c}{$ \Lambda^+_k$} \\
            &  Circle & Square & Triangle & Circle & Square & Triangle \\
        \includegraphics[width=.1\linewidth]{fig/flow/s_00.png} &  
        \includegraphics[width=.1\linewidth]{fig/masks/s_0c.png} &  
        \includegraphics[width=.1\linewidth]{fig/masks/s_0s.png} & 
        \includegraphics[width=.1\linewidth]{fig/masks/s_0t.png} & 
        \includegraphics[width=.1\linewidth]{fig/masks/s_0oc.png} &  
        \includegraphics[width=.1\linewidth]{fig/masks/s_0os.png} & 
        \includegraphics[width=.1\linewidth]{fig/masks/s_0ot.png} \\ 
        \includegraphics[width=.1\linewidth]{fig/flow/s_10.png} &  
        \includegraphics[width=.1\linewidth]{fig/masks/s_1c.png} &  
        \includegraphics[width=.1\linewidth]{fig/masks/s_1s.png} & 
        \includegraphics[width=.1\linewidth]{fig/masks/s_1t.png} & 
        \includegraphics[width=.1\linewidth]{fig/masks/s_1oc.png} &  
        \includegraphics[width=.1\linewidth]{fig/masks/s_1os.png} & 
        \includegraphics[width=.1\linewidth]{fig/masks/s_1ot.png} \\ 
        \includegraphics[width=.1\linewidth]{fig/flow/s_20.png} &  
        \includegraphics[width=.1\linewidth]{fig/masks/s_2c.png} &  
        \includegraphics[width=.1\linewidth]{fig/masks/s_2s.png} & 
        \includegraphics[width=.1\linewidth]{fig/masks/s_2t.png} & 
        \includegraphics[width=.1\linewidth]{fig/masks/s_2oc.png} &  
        \includegraphics[width=.1\linewidth]{fig/masks/s_2os.png} & 
        \includegraphics[width=.1\linewidth]{fig/masks/s_2ot.png} \\ 
       
         \includegraphics[width=.1\linewidth]{fig/masks/b_00.png} &  
        \includegraphics[width=.1\linewidth]{fig/masks/b_0c.png} &  
        \includegraphics[width=.1\linewidth]{fig/masks/b_0s.png} & 
        \includegraphics[width=.1\linewidth]{fig/masks/b_0t.png} & 
        \includegraphics[width=.1\linewidth]{fig/masks/b_0oc.png} &  
        \includegraphics[width=.1\linewidth]{fig/masks/b_0os.png} & 
        \includegraphics[width=.1\linewidth]{fig/masks/b_0ot.png} \\ 
        \includegraphics[width=.1\linewidth]{fig/masks/b_10.png} &  
        \includegraphics[width=.1\linewidth]{fig/masks/b_1c.png} &  
        \includegraphics[width=.1\linewidth]{fig/masks/b_1s.png} & 
        \includegraphics[width=.1\linewidth]{fig/masks/b_1t.png} & 
        \includegraphics[width=.1\linewidth]{fig/masks/b_1oc.png} &  
        \includegraphics[width=.1\linewidth]{fig/masks/b_1os.png} & 
        \includegraphics[width=.1\linewidth]{fig/masks/b_1ot.png} \\ 
        \includegraphics[width=.1\linewidth]{fig/masks/b_20.png} &  
        \includegraphics[width=.1\linewidth]{fig/masks/b_2c.png} &  
        \includegraphics[width=.1\linewidth]{fig/masks/b_2s.png} & 
        \includegraphics[width=.1\linewidth]{fig/masks/b_2t.png} & 
        \includegraphics[width=.1\linewidth]{fig/masks/b_2oc.png} &  
        \includegraphics[width=.1\linewidth]{fig/masks/b_2os.png} & 
        \includegraphics[width=.1\linewidth]{fig/masks/b_2ot.png} \\ 
   
    \end{tabular}
    \captionof{figure}{Flow Capsule inferred object shapes and their visible segmentations masks on Geo and Geo$^+$ test set. In the last row of each dataset where one of the objects is absent, the corresponding mask is empty. }
    \label{fig:geo_masks}
\end{table*}

\begin{table*}[!h]
    \centering
    \begin{tabular}{cccccccc}
    $\image$ & GT & Head &  Torso & Hips & Left Leg 1 & Right Leg 1 & Right Leg 2\\
        \includegraphics[width=.1\linewidth]{fig/masks/e_40.png} &  
        \includegraphics[width=.1\linewidth]{fig/masks/e_4gt.png} & 
        \includegraphics[width=.1\linewidth]{fig/masks/e_4he.png} &  
        \includegraphics[width=.1\linewidth]{fig/masks/e_4t.png} & 
        \includegraphics[width=.1\linewidth]{fig/masks/e_4h.png} & 
        \includegraphics[width=.1\linewidth]{fig/masks/e_4lll.png} & 
        \includegraphics[width=.1\linewidth]{fig/masks/e_4rl.png} & 
        \includegraphics[width=.1\linewidth]{fig/masks/e_4rul.png}\\ 
        \includegraphics[width=.1\linewidth]{fig/masks/e_10.png} &  
        \includegraphics[width=.1\linewidth]{fig/masks/e_1gt.png} &  
        \includegraphics[width=.1\linewidth]{fig/masks/e_1he.png} & 
        \includegraphics[width=.1\linewidth]{fig/masks/e_1t.png} & 
        \includegraphics[width=.1\linewidth]{fig/masks/e_1h.png} & 
        \includegraphics[width=.1\linewidth]{fig/masks/e_1lll.png} &  
        \includegraphics[width=.1\linewidth]{fig/masks/e_1rll.png} & 
        \includegraphics[width=.1\linewidth]{fig/masks/e_1rul.png} \\ 
        \includegraphics[width=.1\linewidth]{fig/masks/e_30.png} &  
        \includegraphics[width=.1\linewidth]{fig/masks/e_3gt.png} & 
        \includegraphics[width=.1\linewidth]{fig/masks/e_3he.png} &  
        \includegraphics[width=.1\linewidth]{fig/masks/e_3t.png} & 
        \includegraphics[width=.1\linewidth]{fig/masks/e_3h.png} & 
        \includegraphics[width=.1\linewidth]{fig/masks/e_3rll.png} &  
        \includegraphics[width=.1\linewidth]{fig/masks/e_3lll.png} & 
        \includegraphics[width=.1\linewidth]{fig/masks/e_3lul.png} \\ 
  
    \end{tabular}
    \captionof{figure}{The ground truth segmentation masks along with sample Flow Capsule visible masks $\Lambda_k^+$ on the exercise test set.}
    \label{fig:ex_masks}
\end{table*}
}

\paragraph{Qualitative analysis on \texttt{Geo} (\Figure{geo_masks})}
Masks shown in Fig.\ \ref{fig:geo_masks} demonstrate that FlowCapsules learns to detect
meaningful part shapes (e.g., a triangle or circle). Indeed, model tends to explain 
a given image in terms of a small number of generic shapes and occlusion of overlapping parts, effectively performing {\em amodal completion} \cite{Hoffman2001}.
This is particularly interesting since the model does not include an explicit regularizer 
that encourages the model to learn a specific number of shapes, or sparsity in the space of shapes.
One might not expect the model to learn to represent the entire shapes where possible 
(e.g.~an entire circle). For example, one might have expected the model to have learned a large number 
of different shapes from which the observed shapes are constructed, especially with occlusion
where the entire shape is often not observed in a single image.
Nevertheless, the model opts to explain the images with relatively few parts, and hence the 
capsule masks tend to cover all the pixels of a shape in the Geo dataset. 
This can be attributed to the architecture we use for mask decoders, and the inductive 
bias of MLPs in generating low-frequency functions~\cite{tancik2020fourier,atzmon2020sal,basri2020frequency,rahaman2019spectral}.
% \at 
% \david{say more about source of inductive bias?}
% \AT{I'll add a sentence referring to the spectral bias in MLPs? SG?}
% \david{That would be great. thanks.}

% Please add the following required packages to your document preamble:
% \usepackage{booktabs}
% \usepackage{multirow}
\begin{table}[t]
\vspace*{-0.1cm}
\begin{tabular}{@{}llccc@{}}
\toprule
                          &           & R-NEM & PSD  & Flow Capsules \\ \midrule
\multirow{4}{*}{Geo}      & Circle    & 0.54 & 0.93 & \textbf{0.94} \\
                          & Square    & 0.56 & 0.82 & \textbf{0.98} \\
                          & Triangle  & 0.58 & 0.90 & \textbf{0.98} \\
                          & All       & 0.56 & 0.88 & \textbf{0.95} \\ \midrule
\multirow{4}{*}{Exercise} & Torso     & 0.32 & 0.57 & \textbf{0.62} \\
                          & Left Leg  & 0.29 & 0.37 & \textbf{0.59} \\
                          & Right Leg & 0.23 & 0.34 & \textbf{0.54} \\
                          & All       & 0.28 & 0.43 & \textbf{0.58} \\ \bottomrule
\end{tabular}
\vspace*{-0.05cm}
\caption{
\textbf{Quantitative / Segmentation -- }
IoU of inferred segment masks w.r.t ground truth on Geo and Exercise data.}
\vspace*{-0.3cm}
\label{tab:iou}
\end{table}

Geo is synthetic, so correct masks for the full shapes are known. 
Since FlowCapsules provide both the part shapes, via $\Lambda_k$, and the
associated visibility masks $\Lambda_k^+$ taking occlusion into account, 
we can compare $\Lambda_k$ to the full ground truth shapes.  
One can then quantify performance using the usual intersection over union (IoU) measure.
FlowCapsules achieves segments with an IoU of \textbf{0.96} on all the shapes, circle, square, 
and triangle  (see Table \ref{tab:iou}). 
This result indicates clearly how well the model encodes the full shape, effectively filling in 
occluded portions of shapes in test images.

\comment{
\todo{Second, comparing this results against Table~\ref{tab:iou} indicates that 
the FlowCapsules layering mechanism is not efficient. 
The model looses accuracy after layering, especially for the circle class.}
\AT{can we drop this sentence? not sure what you are trying to say either}
}

\vspace*{-0.1cm}
\paragraph{Qualitative analysis on \texttt{Exercise} (\Figure{ex_masks})}
On the Exercise dataset, FlowCapsules learn to segment the body into \textit{roughly} rigid parts.
Fig.\ \ref{fig:ex_masks} illustrates the segmentation masks of some of the part capsules.
The masks for individual capsules consistently capture the pixels associated with meaningful 
body parts, such as the head or right leg, regardless of the input image. 
As such, capsule identities are tied to a semantic parts rather than spatial position. 
We also note that the capsules tend to delimit parts at joints, and separate the hips (lower torso) 
from the legs and from the upper torso, even though we do not use a kinematic prior.

% \AT{this seems a repetition w.r.t. PSD section below?}
% \todo{Table~\ref{tab:iou} verifies that Flow Capsule generated segments are more accurate than PSD and R-NEM. 
% Considering that capsule segments are more fine grained than the provided ground truth masks, the quantitative IoUs are significantly better.}

\begin{figure}[t]
\vspace*{-0.2cm}
\centering
\includegraphics[width=.875\linewidth]{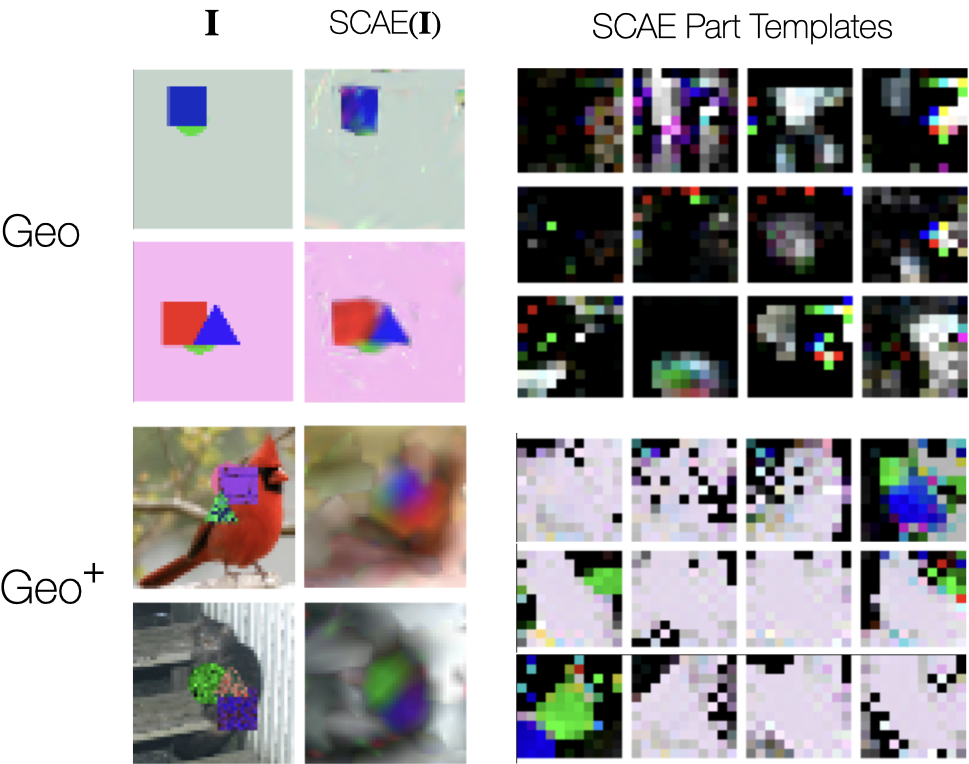}
\caption{(left) SCAE reconstructions after training on Geo and Geo$^+$. 
(right) The learned part templates. 
SCAE approximately reconstructs the image but the part templates are not coherent parts. 
Comparing Geo$^+$ and Geo, the learned parts loose all shape information to enable reconstructing the color, 
texture and background in the images. 
% (cf.\ FlowCapsules part shapes in Fig.\ 
%   \ref{fig:geo_masks}.)
% \AT{should we repeat a column of flowcaps results here?}
}
\vspace*{-0.35cm}
\label{fig:scae}
\end{figure}

\vspace*{-0.1cm}
\paragraph{SCAE (\Figure{scae})}
The most relevant prior work to FlowCapsules vis-a-vis part discovery is SCAE~\cite{kosiorek2019stacked}.
\Figure{scae} shows part templates and image reconstructions generated by SCAE.
Even in simple cases without backgrounds or texture, SCAE fails to segment images into meaningful parts. Unlike FlowCapsules, Fig.~\ref{fig:geo_masks}, it does not have a semantic atomic part definition.
This failure becomes markedly worse for Geo$^+$ when object textures and background are added. FlowCapsules are able to detect and focus on foreground objects with coherent part masks. But SCAE has to reconstruct the background, so the part shapes become general blobs.

\vspace*{-0.1cm}
\paragraph{PSD and R-NEM (\Table{iou})}
We compare the IoU of our masks against PSD and R-NEM~\cite{van2018relational}.
Although PSD additionally receives the \textit{ground truth flow} during training, FlowCapsules consistently outperforms
with equal or better IoUs during testing, on both the Geo and Exercise datasets
(see Tab.\ \ref{tab:iou}).
One difference between PSD and FlowCapsules stems from the way they generate shape masks. 
PSD generates segmentation masks directly using convolutional layers with no encoding of the shape per se. In contrast, FlowCapsules uses a low-dimensional shape code to explicitly model the shape, from which the decoder generates the mask. As such the FlowCapsules encoder disentangles meaningful shape and pose information.
% \AT{we say nothing about R-NEM?}

On Geo$^+$, FlowCapsule IoU performance degrades approximately 10\% to \textbf{0.85} (circle),  
\textbf{0.93} (square),  \textbf{0.90} (triangle) and overall to \textbf{0.89}. 
But compared to results in \Table{iou}, they remains as good or better than 
PSD on the simpler Geo data; we were not able to train PSD effectively on Geo$^+$. 

% Please add the following required packages to your document preamble:
% \usepackage{booktabs}
\begin{table}
\begin{center}
\begin{tabular}{@{}lcccc@{}}
            & \multicolumn{2}{c}{Geo} & \multicolumn{2}{c}{Geo+} \\ \toprule
            & N=4       & N=100         & N=4        & N=100       \\ \midrule
SCAE        & 0.48      & 0.59          & 0.49       & 0.51        \\
FlowCapsule & \textbf{0.79}      & \textbf{0.99}          & \textbf{0.52}       & \textbf{0.74} \\
\bottomrule
\end{tabular}
\end{center}
\vspace{-0.6em}
\caption{
\textbf{Quantitative / Classification: } K-means clustering accuracy with 4 or 100 clusters for Geo and Geo$^+$. 
FlowCapsule part representations yields higher classification accuracy than those learned from SCAE.
% \todo{as opposed to reconstruction based features of SCAE}.
}
\vspace*{-0.35cm}
\label{tab:classification}
\end{table}

\vspace*{-0.1cm}
\subsection{Unsupervised Classification (\Table{classification})}
\vspace*{-0.1cm}

To evaluate FlowCapsules in the broader context of capsule classification, we replace the \textit{primary} 
capsule autoencoder (bottom of the stack) in SCAE \cite{kosiorek2019stacked} with FlowCapsules.
We call the new model \textit{FlowSCAE}.
We then train the top SCAE \textit{object} capsules to reconstruct the pose of FlowCapsules, 
following the original SCAE paper.
We compare the results against SCAE trained on reconstructing images from Geo and Geo$^+$.
SCAE training was modified slightly to produce coloured templates for the GEO dataset, and to 
produce textured templates in the primary capsules for Geo$^+$ (see supplementary material for details).

\Table{classification} reports unsupervised classification results using k-means clustering 
with $N$ clusters, for which the predicted label is set to the most common label in a given cluster.
We report the accuracy with $N{=}4$ and $N{=}100$ clusters.
Note that even though we trained the K-means of FlowSCAE with $N{=}100$ on the Geo data,
the learnt representations contained only $28$ clusters.

\vspace*{-0.1cm}
\subsection{Ablation Studies}
\vspace*{-0.1cm}

To better understand and analyze the significance of our design elements we perform ablations on various parameters. 

\paragraph{Number of capsules~($K$)}
% First we consider robustness to the number of capsules~($K$).
Results in Tab.~\ref{tab:ablation} show that increasing the number of capsules tends to improve IoU performance. 
Given that our model \textit{does not} have an explicit sparsity regularizer on the capsules, this result is intriguing. 
Even with large numbers of capsules available, FlowCapsules does not break shapes into smaller pieces.
Rather, it learns one capsule per shape, relying more heavily on the layer occlusion to explain observed shape variation.
% While the model could have broken shapes into smaller pieces with the greater capacity, 
% Tab.~\ref{tab:ablation} shows that when local parts of a shape move together consistently, 
% the model explains them with a single capsule. 

\paragraph{Encoding length $|s_k|$} 
% We also studied the dependence of IoU performance on the shape encoding length $|s_k|$. 
The models are quite robust Geo and Geo$^+$ data.  
As the encoding dimension decreases from $27$ to $11$, IofU performance changes by only $2\%$. 
Degradation occurs mainly with the circle class, where the circle boundary appears locally linear in places. 
The degradation becomes worse with  $|s_k| =3$, although even then, FlowCapsules still outperforms PSD. 

\paragraph{Number of hidden layers in $\masknet$}
One can hypothesize that deeper decoders can offset issues due to shorter shape encodings. 
% To that end, we vary the number of hidden layers in $\masknet$. 
Table \ref{tab:ablation} shows that increasing decoder depth from $2$ to $6$ improves IoU scores.
With Geo, the deeper decoder produces smoother circles.

\paragraph{Occlusion inductive bias}
Finally, we consider the effect of depth ordering in Eq.~(\ref{eqn:lambdaplus}) for occlusion  handling. 
Without depth ordering, Tab.~\ref{tab:ablation} shows a significant drop in performance.
In this case the masks become smoother and less certain in local regions, and the flow fields appear to be
the result of mixing a larger number of capsules, which tend to fit the observations less well in most cases.

\begin{table}[t]
\begin{tabular}{llll}
\toprule
K  & $|s_k|$  & Geo & Geo$^+$ \\
\midrule
4  & 11 & 0.94 &   0.77\\
8  & 11 & 0.93 &   0.83\\
16 & 11 & 0.94 &   0.88\\
\midrule
8  & 3  & 0.91 &   0.86\\
8  & 27 & 0.96 &   0.89\\ 
\bottomrule
\end{tabular}
\begin{tabular}{lll}
\toprule
Depth  & Decoder   & Geo  \\
\midrule
No & 6-Layer    & 0.54\\
\hline
Yes & 2-Layer   & 0.87\\
Yes & 6-Layer   & 0.96\\
\bottomrule
\end{tabular}
\vspace*{-0.05cm}
\caption{IoU on Geo and Geo$^+$ for different number of capsules, 
encoding lengths, decoder depths, and depth ordering.
% \AT{make two tables same height? e.g. 2,4,6 and Yes/No}
}
\vspace*{-0.1cm}
\label{tab:ablation}
\end{table}

\vspace*{-0.1cm}
\section{Conclusion}
\vspace*{-0.1cm}

We introduce FlowCapsules, an unsupervised method for learning capsule part representations 
(i.e.,~primary capsules). 
The capsule encoder takes as input a single frame and estimates a set of primary 
capsules, each comprising a shape mask in canonical coordinates, a pose transformation 
from canonical to image coordinates, and a scalar representing relative depth.
Training is done in a self-supervised manner from consecutive video frames.
We use a~Siamese architecture to estimate a parametric optical flow field between 
two frames, for which the flow is determined by the poses of corresponding part capsules in the two frames.
% We reiterate that \textit{the capsule encoder only takes one frame as input, and does not require flow at test time}.
Given a single frame, our capsule encoder learns to detect and encode the movable parts in an image. 
This approach differs significantly from other approaches that essentially 
segment the flow field itself into \textit{moving} parts (vs.~\textit{movable} parts in FlowCapsules).

% \AT{david suggested to drop a summary}
% \todo{
% Our model makes the capsule encoder suitable for extracting a part encoding from a 
% single image, and the part encoding disentangles position from shape, which makes 
% them applicable to both segmentation and classification tasks. 
% We evaluate capsule part shape encoding in terms of segmentation and IoU of ground truth masks. 
% We show that our method is not only able to learn an encoding but also able to 
% generate more precise masks than in previous work.
% We further evaluate the position encoding of our capsules by replacing the 
% primary encoder of SCAE~\cite{kosiorek2019stacked}.
% The new, unsupervised object classification model, FlowSCAE,  outperforms SCAE, 
% especially when the complexity of images increase~(background and object texture).}
%
% \paragraph{Future work}

Empirical results show that motion self-supervision in FlowCapsules is effective on real and synthetic data, 
learning meaningful representations, completing shapes when partially occluded.
While formulated and tested within a specific capsule framework, our approach to self-supervised parts discovery 
is applicable to myriad of encoder architectures, and to other approaches that currently use an image-reconstruction 
loss or rely on optical flow as input. Combining motion-based self-supervision with attention-based encoders  
\cite{locatello2020object} would enhance compositoinality, allowing scenes with different numbers of objects.
Future work will also include scaling to larger video datasets and 3D parts.
To that end it will be important to extend the approach to include camera motion, and to handle
large motions of small objects for which more sophisticated losses for self-supervised learning will be necessary.
Alternatively, the FlowCapsules framework should be directly applicable to 3D observations, like point cloud data
\cite{zhao20193d}.

% One way to do so would be to use off the shelf optical flow estimators for pre-training FlowCapsules. 
% Given recent advances in the unsupervised flow estimation literature, e.g., incorporating different 
% degrees of smoothness and accounting for the edges, we hope to further improve our part encoding. 
% Another interesting direction would be to model the flow, masks and transformations in 3D using point clouds. 
% A 3D setup would make natural image understanding easier for the model, especially 
% with change of perspective and camera movement.

% \AT{shouldn't be ``straight'' and say something about \textit{panoptic} segmentation? Currently if \textit{multiple} objects appear...}

% Acknowledgements should only appear in the accepted version.
\section*{Acknowledgements}
We thank Luca Prasso and Deqing Sun for help preparing datasets, Dirk Weissenborn and Jakob Uszkoreit for helpful discussions in the initial stages of the project. and Zhenjia Xu for help with the PSD experiment setup.

\bibliography{main.bib}
\bibliographystyle{icml2020}

%%%%%%%%%%%%%%%%%%%%%%%%%%%%%%%%%%%%%%%%%%%%%%%%%%%%%%%%%%%%%%%%%%%%%%%%%%%%%%%
%%%%%%%%%%%%%%%%%%%%%%%%%%%%%%%%%%%%%%%%%%%%%%%%%%%%%%%%%%%%%%%%%%%%%%%%%%%%%%%
% DELETE THIS PART. DO NOT PLACE CONTENT AFTER THE REFERENCES!
%%%%%%%%%%%%%%%%%%%%%%%%%%%%%%%%%%%%%%%%%%%%%%%%%%%%%%%%%%%%%%%%%%%%%%%%%%%%%%%
%%%%%%%%%%%%%%%%%%%%%%%%%%%%%%%%%%%%%%%%%%%%%%%%%%%%%%%%%%%%%%%%%%%%%%%%%%%%%%%
\clearpage

}
\section{Supplemantary Material}
\subsection{SCAE Training Details}
While comparing FlowCapsules against SCAE, we updated SCAE training at various spots to make it more
suitable for Geo and Geo$^+$ datasets. Here we detail these changes. First, We resized input images to
$48 \times 48$ for memory reasons. Second,
we added the option of inferring the background color as well as background image using a two level MLP.
Similarly, we added the option of adding color or texture to each template. To enable colorization and texturization based on input image, the primary capsule features are passed to the template decoder. The color/texture is generated by a 2 layer MLP (32 dimensional hidden representation). The original fixed templates are used as masks and multiplied to the output of the color/texture MLP. 

For generating a background template, we use the second to last hidden representation of the primary encoder as the image embedding. We pass the image embedding through a 2 layer MLP (32 dimensional hidden representation). We mix this background template with a presence probability of $0.5$.

All the other parameters, including training schedule is kept the same as the original SCAE.

\subsection{Exercise masks}

\begin{figure}[h]
    \centering
    \includegraphics[width=0.975\linewidth]{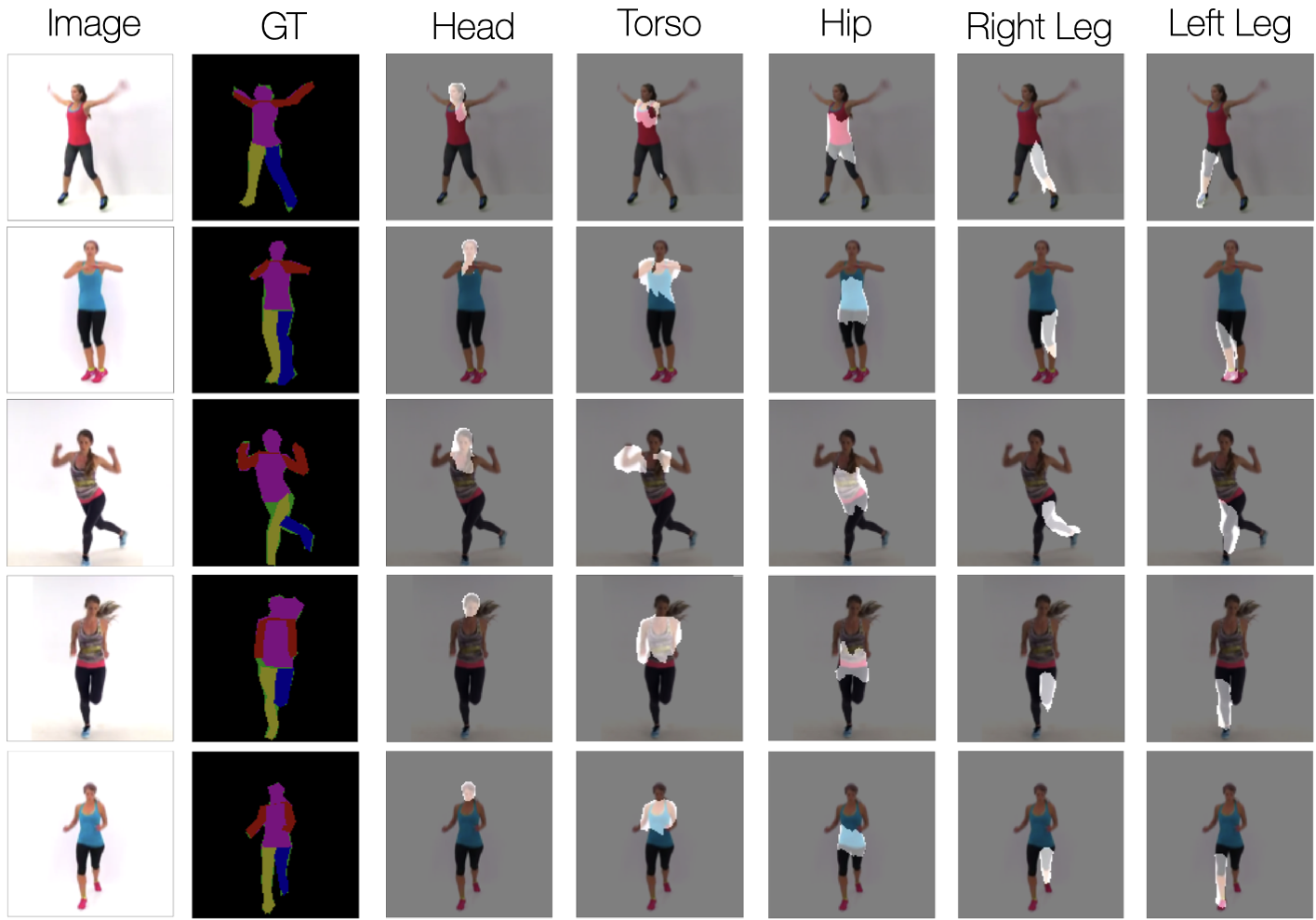}
    \vspace*{-0.1cm}
    \caption{The ground truth segment masks along with sample FlowCapsule masks $\Lambda_k^+$ on Exercise test data.}
    \label{fig:ex_masks}
\end{figure}

%\end{document}

\end{document}

% This document was modified from the file originally made available by
% Pat Langley and Andrea Danyluk for ICML-2K. This version was created
% by Iain Murray in 2018, and modified by Alexandre Bouchard in
% 2019 and 2020. Previous contributors include Dan Roy, Lise Getoor and Tobias
% Scheffer, which was slightly modified from the 2010 version by
% Thorsten Joachims & Johannes Fuernkranz, slightly modified from the
% 2009 version by Kiri Wagstaff and Sam Roweis's 2008 version, which is
% slightly modified from Prasad Tadepalli's 2007 version which is a
% lightly changed version of the previous year's version by Andrew
% Moore, which was in turn edited from those of Kristian Kersting and
% Codrina Lauth. Alex Smola contributed to the algorithmic style files.